\documentclass[10pt,twocolumn,letterpaper]{article}

\usepackage[pagenumbers]{cvpr} % To force page numbers, e.g. for an arXiv version

\usepackage[dvipsnames]{xcolor}

\usepackage{graphicx}
\usepackage{amsmath, bm}
\usepackage{amssymb}
\usepackage{booktabs}

\usepackage{algorithm,algpseudocode}
\usepackage{algorithmicx}
\usepackage{xcolor}         % colors

\usepackage[font=small]{caption} % caption
\usepackage{ulem} % 删除线
\usepackage{bbm} 
\usepackage{multirow} 
\usepackage{tabularx}
\usepackage{longtable}
\usepackage{arydshln}
\usepackage{array}    % For more control over table column specifications

\newcommand{\AtIns}{\mathcal{A}_{t, \text{ins}}}

\definecolor{cvprblue}{rgb}{0.21,0.49,0.74}
\usepackage[pagebackref,breaklinks,colorlinks,citecolor=cvprblue]{hyperref}
\usepackage[capitalize]{cleveref}

\crefname{section}{Sec.}{Secs.}
\Crefname{section}{Section}{Sections}
\Crefname{table}{Table}{Tables}
\crefname{table}{Tab.}{Tabs.}

\def\paperID{*****} % *** Enter the Paper ID here
\def\confName{CVPR}
\def\confYear{2024}

\title{Focus on Your Instruction: Fine-grained and Multi-instruction \\ Image Editing by Attention Modulation}

\author{
\vspace{0.2cm}
Qin~Guo$^{1,2}$, Tianwei~Lin$^2$ \\
$^1$Peking University, $^2$Horizon Robotics \vspace{0.2cm} \\
{\tt\small guoqin@stu.pku.edu.cn, tianwei.lin@horizon.cc}
}

\begin{document}
% \maketitle
\twocolumn[{
\maketitle
\centering
    \begin{tabular}{ccc}
    \multicolumn{3}{c}{\includegraphics[width=0.85\linewidth]{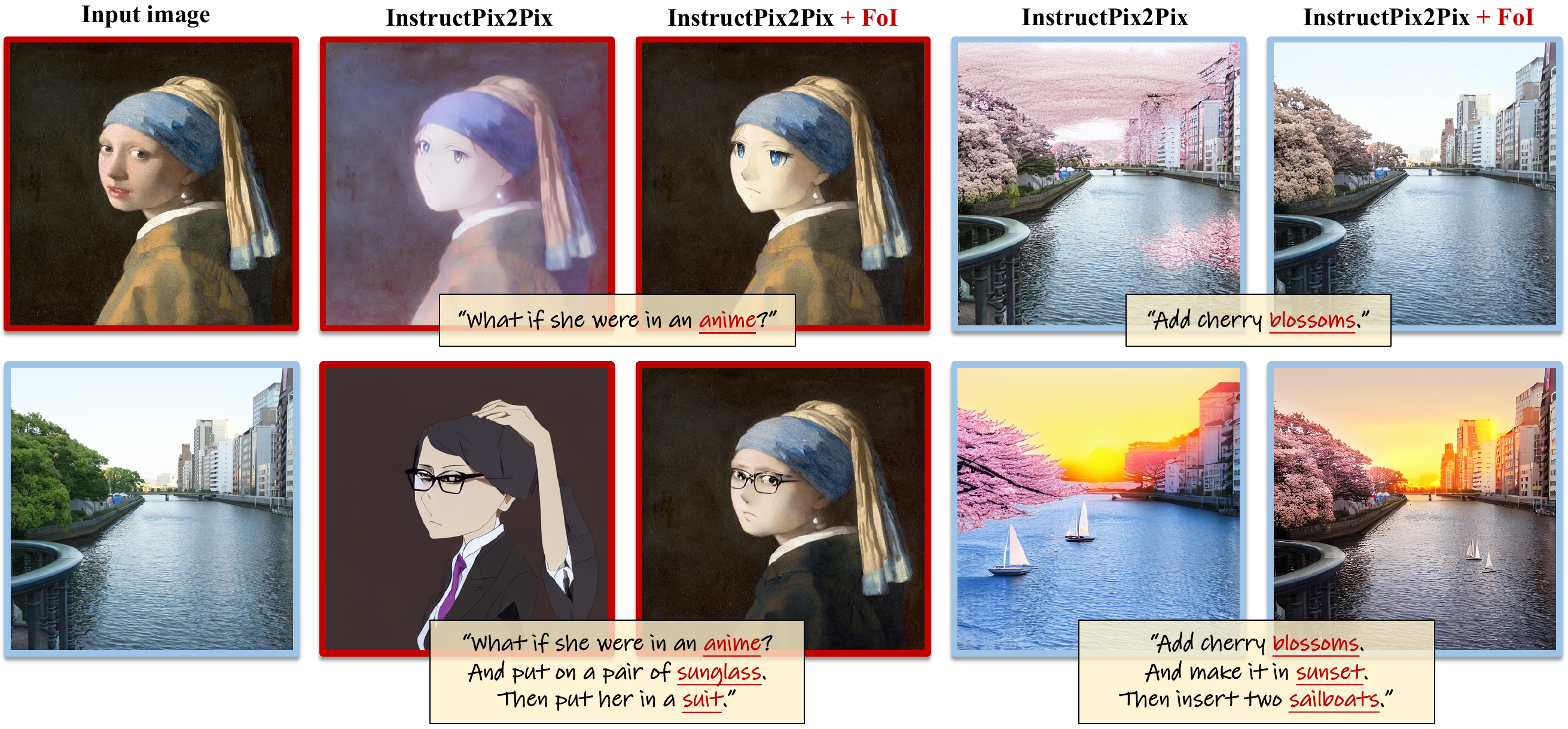}} \\
    \end{tabular}
  \captionof{figure}{
Models like InstructPix2Pix (IP2P)~\cite{instructpix2pix} can edit images with given instruction.
Yet, they face challenges like over-editing and wrong editing areas, especially with multi-instruction.
Our FoI utilizes inherent grounding capability of IP2P to identify precise editing regions, then focuses  on them, enabling effective editing. Notably, FoI does not require extra training or test-time optimization.
}
  \label{fig:teaser}
 \vspace*{0.4cm}
}]

\begin{abstract}
Recently, diffusion-based methods, like InstructPix2Pix (IP2P), have achieved effective instruction-based image editing, requiring only natural language instructions from the user.
However, these methods often inadvertently alter unintended areas and struggle with multi-instruction editing, resulting in compromised outcomes. 
To address these issues, we introduce the \textbf{Focus on Your Instruction (FoI)}, a method designed to ensure precise and harmonious editing across multiple instructions without extra training or test-time optimization.
In the FoI, we primarily emphasize two aspects: (1) precisely extracting regions of interest for each instruction and (2) guiding the denoising process to concentrate within these regions of interest.
For the first objective, we identify the implicit grounding capability of IP2P  from the cross-attention between instruction and  image, then develop an effective mask extraction method.
For the second objective, we introduce a cross attention modulation module for rough  isolation of target editing regions and unrelated regions. Additionally, we introduce a mask-guided disentangle sampling strategy to further ensure clear region isolation.
Experimental results demonstrate that FoI surpasses existing methods in both quantitative and qualitative evaluations, especially excelling in multi-instruction editing task.
The code is available at \url{https://github.com/guoqincode/Focus-on-Your-Instruction}.
\end{abstract}    
\section{Introduction}
\label{sec:intro}

Large-scale Text-to-Image (T2I) diffusion models~\cite{latentdiffusion,Imagen,sdxl,ramesh2022hierarchical,balaji2022ediffi,xue2023raphael,nichol2021glide,gal2022image,ruiz2023dreambooth} have achieved remarkable diversity and realism in image generation, garnering widespread attention. Trained on extensive image-text datasets~\cite{schuhmann2022laion}, these advanced T2I models excel in various generation tasks. However, their direct application to image editing is limited, often lacking the necessary precision for controlling specific objects or attributes within images.

When editing images, a visual creator typically begins by identifying the regions to be edited and then focuses on modifying these regions. For multiple edits, ensuring the collective result is cohesive is crucial. Despite recent remarkable advances in text-based image editing~\cite{sdedit,imagic,hertz2022prompt,nulltext,wallace2023edict,pnp,parmar2023zero,instructpix2pix,diffedit,nichol2021glide,zhang2023adding}, the precisely editing of targeted areas without affecting unrelated regions remains a significant challenge. These methods often struggle to accurately pinpoint the editing areas, leading to unintended modifications in non-targeted areas and resulting in suboptimal outcomes. Furthermore, they typically struggle to simultaneously execute edits in multiple directions,  further limiting their utility in complex editing tasks.

\begin{figure}[t!]
    \centering
    \begin{tabular}{c}
    \includegraphics[width=0.9\linewidth]{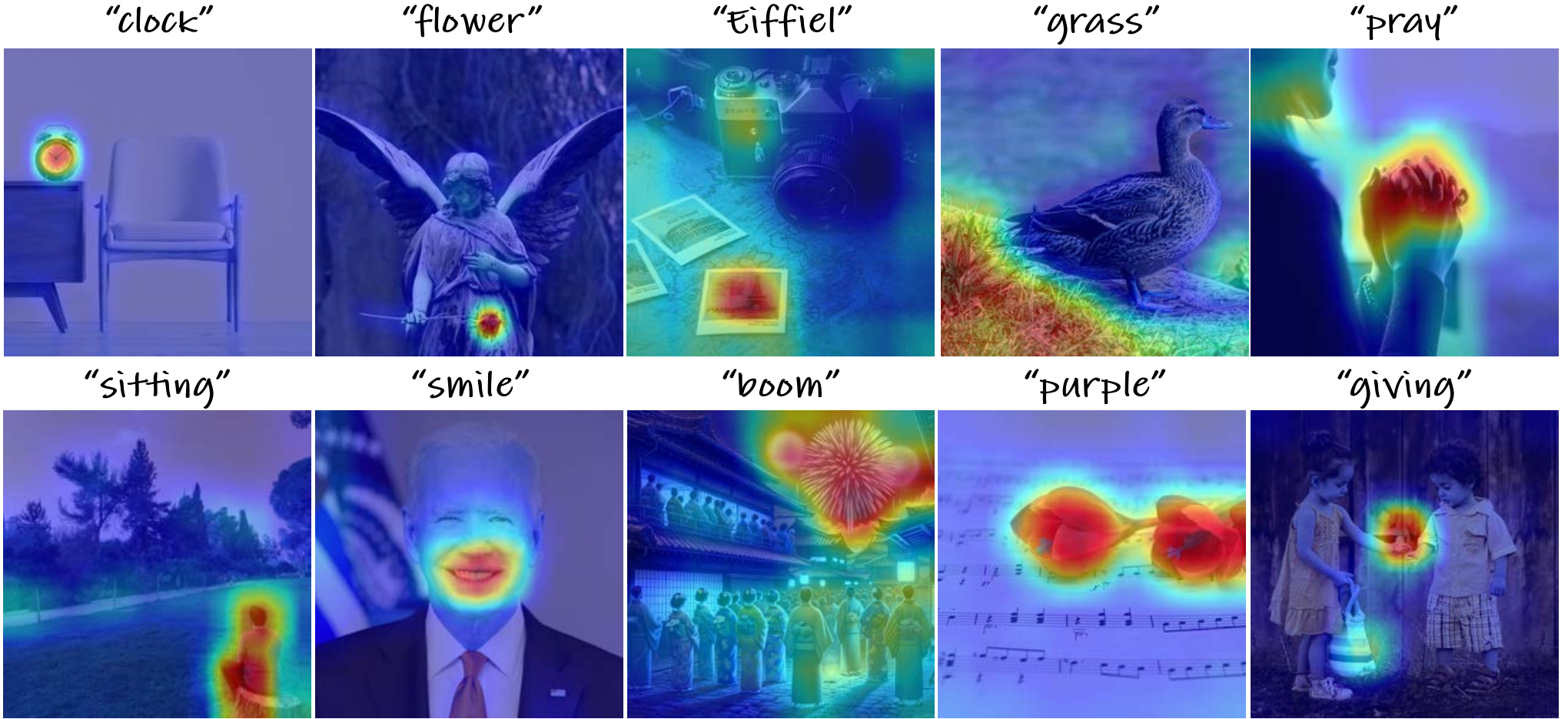} \\
    \end{tabular}
    \caption{
Visualization of cross-attention maps \textbf{in the initial denoising step} illustrates the fine-grained implicit grounding capability of IP2P~\cite{instructpix2pix} for \textbf{nouns}, as well as \textbf{verbs} and \textbf{adjectives}.
    }
    % \vspace{-16pt}
    \vspace{-8pt}
  \label{fig:grounding}
\end{figure}

\begin{figure}[t!]
    \centering
    \begin{tabular}{c}
    \includegraphics[width=0.97\linewidth]{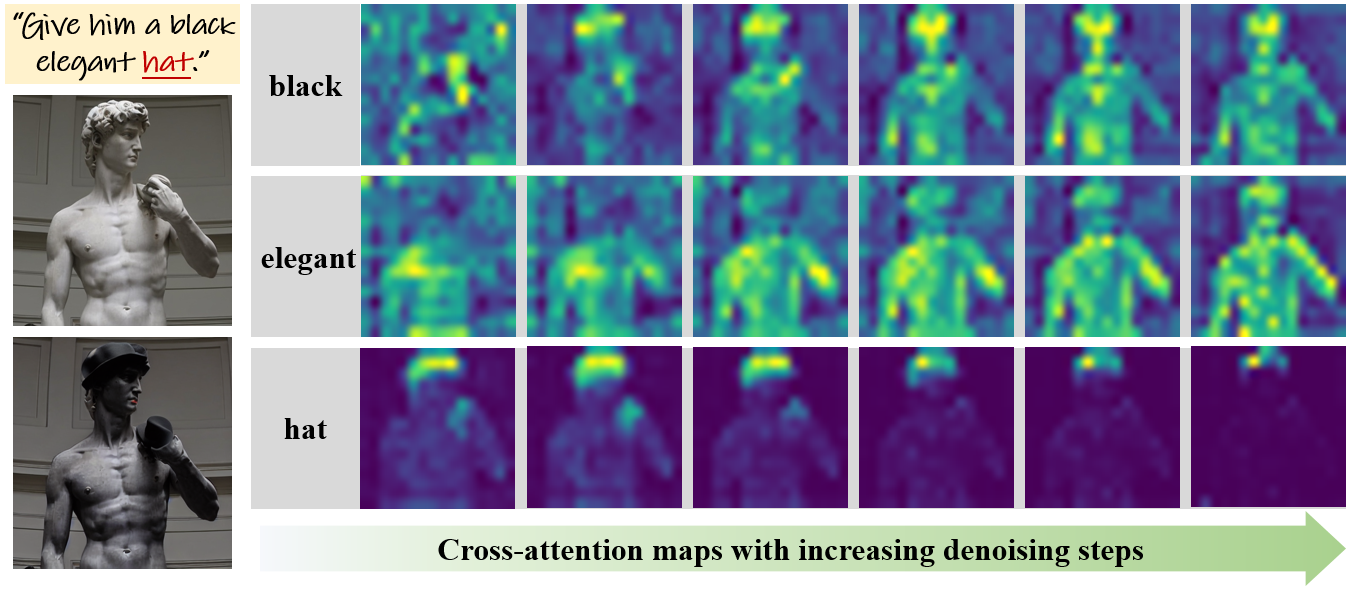} \\
    \end{tabular}
    \caption{
  Visualization of cross attention maps obtained from IP2P~\cite{instructpix2pix}, associated with different words. 
  Two key observations: (a) the placement of \textit{``hat''} is accurately identified early on, and (b) the attention maps for adjectives \textit{``black''} and \textit{``elegant''} are excessively disperse, leading to over-editing.
  }
  % \vspace{-18pt}
  \label{fig:concept}
\end{figure}

IP2P~\cite{instructpix2pix} offers an intuitive and fidelity-preserving approach for instruction-based image editing, bypassing the need for extensive descriptions of input and output images. 
However, as shown in~\cref{fig:teaser}, IP2P has a propensity for over-editing, which is also indicated in recent studies~\cite{mirzaei2023watch,joseph2023iterative}.
In our analysis of IP2P, we unveil its powerful implicit grounding ability developed through training on a synthetic pairwise dataset. As shown in~\cref{fig:grounding}, in the cross-attention map of initial denoising step, we can observe precise alignment between keywords and their spatial locations in the image. This  effective grounding extends to even adjectives and verbs. This sharply contrasts with the evolving attention maps in models like Stable Diffusion~\cite{hertz2022prompt,daam,attend,Astar}. However, as depicted in~\cref{fig:concept}, while IP2P effectively locates items like a ``hat", other instruction words may inadvertently affect unrelated areas, leading to unintended edits. 
To our knowledge, no existing methods have harnessed IP2P's potent grounding ability to enhance its editing ability.

To address the limitations of current image editing methods and align with the editing paradigm of visual creators, we introduce \textbf{Focus on Your Instruction (FoI)},
a method developed atop the IP2P framework.
FoI is specifically designed for precise and harmonious multi-instruction editing, and notably, it achieves this without requiring additional training or test-time optimization.
\textbf{First}, we utilize IP2P's implicit grounding ability to identify the areas of interest for each instruction. \textbf{Then}, we introduce cross-condition attention modulation, leveraging null-instruction cross-attention to modulate the cross-attention calculation with instruction, focusing each instruction on its corresponding area and implicitly reducing interference between different instructions. \textbf{Finally}, we propose a mask-guided disentangle sampling strategy, aimed at accurately separating editing and non-editing regions, disentangling the overall editing direction from preserving the original image's direction, and enhancing the model's robustness in hyperparameter selection. Experimental results demonstrate that FoI outperforms existing methods in both quantitative and qualitative evaluations, especially in multi-instruction editing tasks.

Our contributions can be summarized as follows:

\begin{itemize}
    \item We introduce FoI, a method that leverages the grounding ability of IP2P for precise and harmonious multi-instruction editing, without the need for extra training or test-time optimization.
    \item We propose cross-condition attention modulation to ensure each instruction is focused on its corresponding area, thereby reducing interference. This method employs cross-attention with null-instruction to modulate the cross-attention calculation with instruction.
    \item Development of a mask-guided disentangle sampling strategy, isolating editing regions and distinguishing between editing and preserving directions.
    \item Demonstrated excellence of FoI in experiments, outperforming existing methods quantitatively and qualitatively, particularly in multi-instruction editing tasks.
\end{itemize}

\section{Related Work}
\label{sec:relate}
\noindent\textbf{Text-guided Image Editing.} 
Early methods mainly relied on Generative Adversarial Networks (GANs)~\cite{GANs,styleclip,StyleGAN_NADA,deltaedit,xu2022predict}, excelling in specific domains like faces and flowers, but with limited generality. 
Recently, methods based on diffusion models~\cite{sohl2015deep,ho2020denoising} have showcased unprecedented prowess in image generation and editing~\cite{latentdiffusion,Imagen,sdxl,ramesh2022hierarchical,balaji2022ediffi,xue2023raphael,nichol2021glide}.
SDEdit~\cite{sdedit} leverages these models in a two-step process of noise addition and denoising to align with prompts.
Imagic~\cite{imagic} fine-tunes the diffusion model for each image, focusing on generating variants for objects. Prompt2Prompt(P2P)~\cite{hertz2022prompt} and PnP~\cite{pnp} explore attention and feature injection for improving image editing performance. 
Compared to P2P, PnP can directly edit real images. 
To adapt P2P for real image editing, Null-Text Inversion (NTI)~\cite{nulltext} proposes updating the null text embedding for precise reconstruction and editing~\cite{ho2022classifier}.
Blended Diffusion~\cite{blend_diffusion,blend_diffusion_latent} achieves local editing using user-designed masks and prompts.
IP2P~\cite{instructpix2pix} streamlines image editing by directly applying instructions, removing the need for detailed descriptions or masks. 
This approach not only bypasses reconstruction flaws in inversion-based methods~\cite{DDIM,hertz2022prompt,diffedit} but also avoids lengthy test-time optimization~\cite{nulltext,wallace2023edict,parmar2023zero,pnp}, enhancing image fidelity.

\noindent\textbf{Locating the Targeted Editing Area.} 
Precise editing areas localization is crucial to prevent unintended image changes. 
Text2Live~\cite{text2live} utilizes CLIP~\cite{clip} for optimizing additive image layers. FEAT~\cite{hou2022feat} and CoralStyleCLIP~\cite{revanur2023coralstyleclip} leverage StyleGAN's latent codes for domain-specific local editing. Diffedit~\cite{diffedit} and Watch Your Steps~\cite{mirzaei2023watch} generate masks by contrasting different noise predictions. InstructEdit~\cite{instructedit} and OIR~\cite{objectaware} use text-conditioned segmentation models~\cite{grounding_dino,sam} for identifying \textbf{\textit{ existing objects }} specified for editing but struggle with fine-grained editing. 
LPM~\cite{patashnik2023localizing} clusters self-attention maps to pinpoint objects based on cross-attention values, primarily focusing on object-level shape variations. However, most open domain visual editing works face challenges in detailed editing and preserving the original image's fidelity. 
For instance, with an instruction like \textit{``put a Disney headband on her.''}, the aim is to simply add the headband, yet typical methods often alter identity features or other image areas.
By leveraging IP2P's implicit grounding ability, our method accurately targets the most relevant areas for each instruction, achieving finer granularity than prior methods, with only a minimal increase in computational overhead.

\noindent\textbf{Multi-instruction Image Editing.} 
A key challenge involves effectively guiding models to target specific editing areas for each instruction, while minimizing interference among instructions to ensure harmonious multi-instruction outcomes.
Recent work such as EMILIE~\cite{joseph2023iterative} focuses on iterative multi-instruction editing but overlooks IP2P's over-editing issues, mainly addressing image quality decline in successive edits. 
Existing instruction-based methods~\cite{instructpix2pix,instruct_guiding,instructdiffusion,zhang2023magicbrush,hive} often struggle with multi-instruction tasks.
Contemporary methods~\cite{dpl,objectaware} in multi-object editing, using inversion-based techniques~\cite{DDIM}, focus mainly on object-level replacement. These methods, requiring complex optimization, struggle with fine-grained editing and tend to be time-consuming.
In contrast, our approach avoids additional training or test-time optimization, and can be easily integrated with existing instruction-based models.

\begin{figure*}[!t]
    \centering
    \begin{tabular}{c}
    \includegraphics[width=1.0\linewidth]{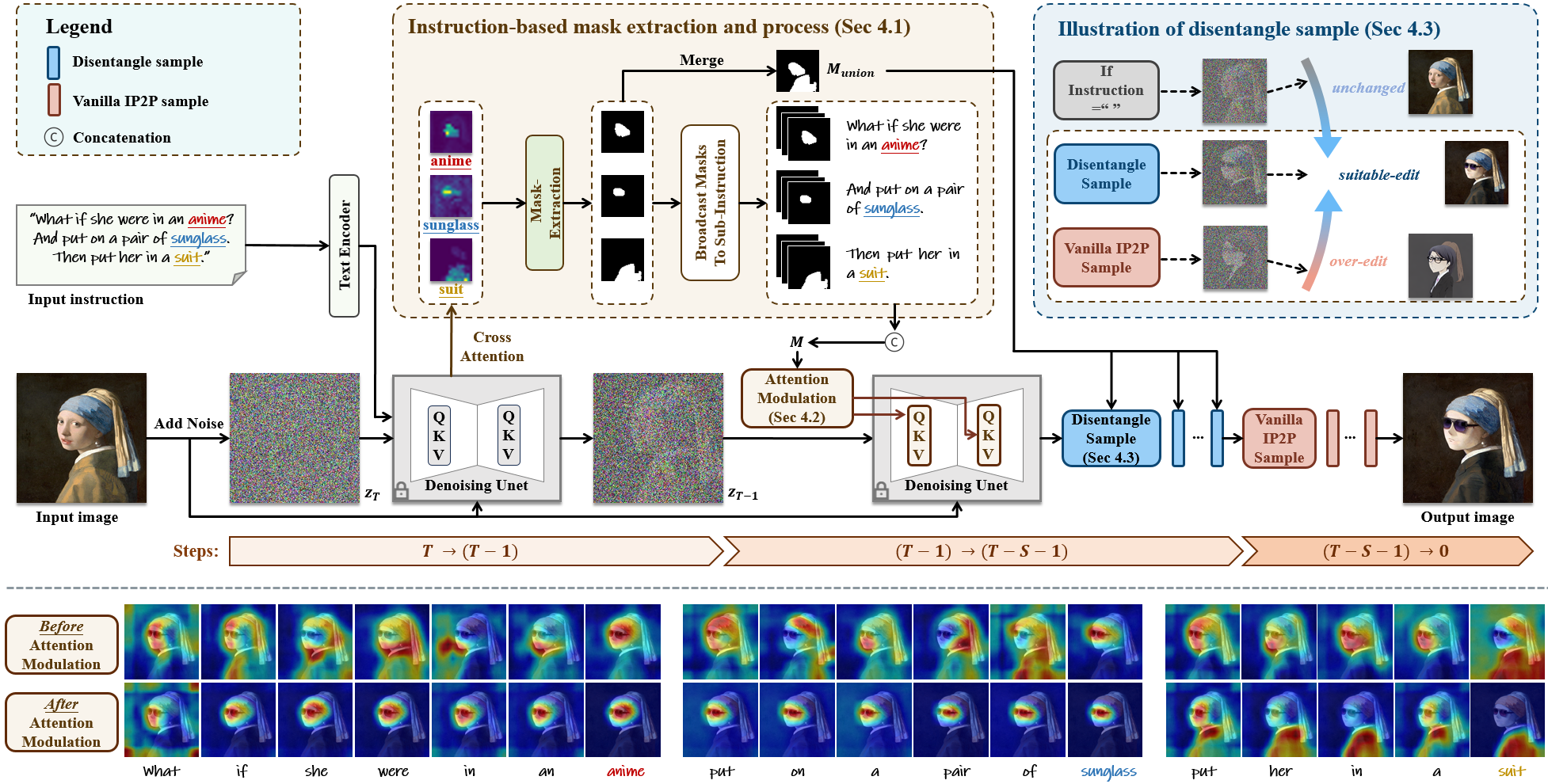} \\
    \end{tabular}
    \vspace{-8pt}
  \caption{
  \textbf{Framework of FoI.} FoI is designed to perform precise single-instruction edits and coordinated multi-instruction edits, all within a single forward pass. Firstly, a unique mask for each sub-instruction is extracted at the start of the denoising step, as described in~\cref{sec:get_mask}. Next, we use cross-condition attention modulation to focus each instruction on its interest area and reduce interference, elaborated in~\cref{sec:attention_modulation}(the bottom figure illustrates the cross-attention map before and after attention modulation). Finally, a disentangle sample method isolates editing areas, detailed in~\cref{sec:disentangle_sample}.
  }
  \vspace{-8pt}
  \label{fig:method}
\end{figure*}
\section{Preliminaries}
\label{sec:prelim}
\noindent\textbf{InstructPix2Pix.}
Given an image \(I\), IP2P edits it following given editing instruction \(T\). 
IP2P undergoes supervised training on a dataset synthesized using P2P~\cite{hertz2022prompt} and GPT-3~\cite{brown2020language}. Each entry in the dataset includes the original image \(I\), the editing instruction \(T\), and the corresponding edited result \(I_\text{e}\).
IP2P is constructed upon on the Stable Diffusion framework~\cite{latentdiffusion}, employing a VQ-VAE~\cite{van2017neural} with an encoder \(\mathcal{E}\) and a decoder \(\mathcal{D}\) to enhance efficiency and quality. 
For training, noise \(\epsilon \sim \mathcal{N}(0, 1)\) is added to \(z = \mathcal{E}(I_\text{e})\) to create noisy latent \(z_t\), with the noise level set by a random timestep \(t \in T\). The denoiser, \(\epsilon_\theta\), initially with Stable Diffusion weights~\cite{latentdiffusion}, is fine-tuned to minimize \(\mathbb{E}_{I_\text{e}, I, \epsilon, t} \big[\Vert \epsilon - \epsilon_\theta (z_t, t, I, T) \Vert_2^2\big]\). Conditions are intermittently omitted during training~\cite{liu2022compositional} by setting \(I = \emptyset_I\) or \(T = \emptyset_T\). The vanilla IP2P score estimate is as follows:
% \vspace{-4pt}
\begin{align}
    \label{eq:modified.score.estimate}
    \tilde{\epsilon}_\theta (&z_t, t, I, T) = \textcolor{teal}{\epsilon_\theta(z_t, t, \emptyset_I, \emptyset_T)} \notag \\ 
                                & + s_I \big( \textcolor{blue}{\epsilon_\theta(z_t, t, I, \emptyset_T)} - \textcolor{teal}{\epsilon_\theta(z_t, t, \emptyset_I, \emptyset_T)}\big) \notag \\
                                & + s_T \big(\textcolor{red}{\epsilon_\theta(z_t, t, I, T)} - \textcolor{blue}{\epsilon_\theta(z_t, t, I, \emptyset_T)}\big)
\end{align}
Studies~\cite{instructpix2pix,mirzaei2023watch,haque2023instruct} highlight the importance of balancing image guidance \(s_I\) and text guidance \(s_T\). An increase in \(s_I\) preserves image details but reduces the impact of instructions, while an increase in \(s_T\)  poses risks of over-editing.
Consequently, \( \textcolor{blue}{\epsilon_\theta(z_t, t, I, \emptyset_T)} \) estimates scores for image preservation, and \(\textcolor{red}{\epsilon_\theta(z_t, t, I, T)}\) for applying edits.

\noindent\textbf{Cross Attention in InstructPix2Pix.}
IP2P incorporates textual features in image editing through a cross-attention mechanism~\cite{attention}.
This process generates cross-attention maps $\mathcal{A}_t \in \mathbb{R}^{r \times r \times N}$ at each denoising step $t$ for every token ($N$ tokens tokenized using CLIP~\cite{clip}'s tokenizer) in the input instruction, where $r \in \{64, 32, 16, 8\}$~\cite{hertz2022prompt,attend,Astar}.
Because IP2P integrates the original image into the input channels of its U-Net, the behavior of its attention mechanism exhibits distinctions compared to Stable Diffusion~\cite{latentdiffusion}. We denote the cross attention map in \(\textcolor{red}{\epsilon_\theta(z_t, t, I, T)}\) as $\AtIns$.
\section{Method}
\label{sec:method}

Given the input image \(I\) and the composite  instruction \(T\), composed of \(k\) sub-instructions \(\{T_1, T_2, \ldots, T_k\}\), our goal is to edit \(I\) with \textbf{(1)} precise execution of each sub-instruction in \(T\), and \textbf{(2)} harmonious execution of \(T\) as a whole. 
We believe the core challenge here is \textbf{\textit{how to precisely directing instructions towards their corresponding areas of interest.}}
To solve this challenge, we propose FoI, with overall framework illustrated in~\cref{fig:method}.
In this section, we first discuss how to find precise area of interest for each sub-instruction (\cref{sec:get_mask}).
Then, we introduce how to guide the denoising process to proper direction where each instruction focus on its own interest area, with cross-attention modulation (\cref{sec:attention_modulation}) and disentangle sampling strategy (\cref{sec:disentangle_sample}).

% \vspace{-4pt}
\subsection{Extracting Instruction-Based Masks}
\label{sec:get_mask}

Inspired by the segmentation capabilities of large-scale diffusion models~\cite{tian2023diffuse, wu2023diffumask, wu2023datasetdm, karazija2023diffusion}, our analysis of IP2P uncovers its precise location-finding ability in early denoising steps, evident from cross-attention maps in~\cref{fig:grounding}. Demonstrated in~\cref{fig:concept}, IP2P quickly identifies where objects, like ``hat'' should be placed. We harness this robust grounding capability to extract areas of interest for each instruction from IP2P's cross-attention maps.

Previous studies~\cite{hertz2022prompt,attend,Astar,xie2023boxdiff} have shown that attention maps with a resolution of $16 \times 16$ capture the most detailed semantic information. Accordingly, we use attention maps with a resolution of \( r = 16 \) for extracting masks.

In each sub-instruction \(T_i\), we identify a keyword \(e_i\), which represents either the target object for editing, an object to be added, or an object inferred from the context, as specified in the sub-instruction.
We begin by applying a Gaussian filter~\cite{attend} to the corresponding cross-attention map $\mathcal{A}_{t}[\bm{e_i}] \in \mathbb{R}^{r \times r}$. This step ensures that each patch in the map is a linear combination of its neighboring patches in the original map.
We then use a direct and effective algorithm for mask extraction, enhancing the cross-attention map \( \mathcal{A}_{t}[\bm{e_i}] \) iteratively. The algorithm operates through a sequence of operations, repeated \(\gamma\) times. In each cycle, the map is squared and then normalized via min-max scaling to the $[0,1]$ range. This iterative approach is designed to incrementally heighten the contrast between target regions and surrounding areas, as detailed in the following equation:
\begin{equation}
    \mathcal{A}_{t}[\bm{e_i}] = \underbrace{\text{norm} \left( \text{norm} \left( \cdots \text{norm} \left( \mathcal{A}_{t}[\bm{e_i}]^2 \right) \cdots \right)^2 \right)^2}_{\gamma \text{ times}}
\end{equation}
Here, norm denotes the min-max normalization process, scaling the values within the map to a [0,1] range.
Upon completing the \(\gamma\) iterations, we apply a threshold \( \tau \) to compute the mask $\mathcal{M}_{e_i} = \mathbbm{1}(\mathcal{A}_{t}[\bm{e_i}] \geq \tau)$. This mask, denoting the area of interest for the \(i\)-th sub-instruction, has dimensions \(\in \mathbb{R}^{r \times r}\).
~\cref{fig:method} displays the effective results of our method in extracting masks for each sub-instruction.

\subsection{Cross Condition Attention Modulation}
\label{sec:attention_modulation}
For fine-grained editing, confining each instruction within its mask is essential. We introduce the cross-condition attention modulation. This method utilizes the cross-attention map with null-instruction to modulate the cross-attention calculation with instruction, thereby reducing the impact of instruction on irrelevant areas and decreasing interference between different instructions when multiple instructions are present.
To be specific, we preserve the masked region's attention in the computation of \(\textcolor{red}{\epsilon_\theta(z_t, t, I, T)}\), while substituting attention in the external regions with that from \(\textcolor{blue}{\epsilon_\theta(z_t, t, I, \emptyset_T)} \). The modified cross attention function is defined as:
\begin{equation}
\label{eq:modulating_attention}
\mathcal{A'}_{t,ins} = \text{softmax}\left(\frac{(\mathcal{X} + \Delta\mathcal{X}) \odot \mathcal{M} + \mathcal{Y} \odot (1-\mathcal{M})}{\sqrt{d}}\right)
\end{equation}
Here, \(d\) represents the latent projection dimension. The terms are defined as follows:
\begin{align}
\mathcal{X} &= Q_{I,T}K^T_{I,T}, \\
\mathcal{Y} &= Q_{I,\emptyset_T}K^T_{I,\emptyset_T}, \\
\Delta\mathcal{X} &= \boldsymbol{\alpha} \odot \xi(t) \odot (\max(Q_{I,T}K^T_{I,T}) - Q_{I,T}K^T_{I,T}) \label{eq:DeltaX}
\end{align}
where \(Q_{I,T}\) and \(K_{I,T}\) are the query and Key in \(\textcolor{red}{\epsilon_\theta(z_t, t, I, T)}\) respectively, and \(Q_{I,\emptyset_T}\) and \(K_{I,\emptyset_T}\) are the query and key in \(\textcolor{blue}{\epsilon_\theta(z_t, t, I, \emptyset_T)} \) respectively.
The attention mask \(\mathcal{M}\) is constructed by initially \textbf{\textit{broadcasting}} the mask \(\mathcal{M}_{e_i}\) of each keyword across its corresponding sub-instruction \(T_i\). Subsequently, these broadcasted masks are concatenated for all sub-instructions, resulting in an initial dimension of \(\mathcal{M}\) being $\mathbb{R}^{(r \times r) \times N}$. This mask is then adaptively interpolated across each cross-attention layer.

We employ \(\Delta\mathcal{X}\) to subtly enhance attention values within the mask. This allows for precise control over the relative strengths of different sub-instructions, enabling fine-grained control over the intensity of each sub-instruction.
This is achieved by selectively adjusting the values in the coefficient vector \(\boldsymbol{\alpha}\), which is initially set to all ones. Through strategic modifications to specific values within \(\boldsymbol{\alpha}\), we can finely tune the intensity of each sub-instruction.
This method directs attention within the mask and enables flexible adjustment of each sub-instruction's relative intensity, ensuring focused and controlled effects during the editing process.

In~\cref{eq:DeltaX}, the timestep-related weight term is:
\begin{equation}
    \xi(t) = 0.05 * t^4
\end{equation}
and the timestep $t \in [0, 1]$ has been normalized.

After mask extraction, we apply cross-condition attention modulation across all remaining denoising steps.
The bottom part of~\cref{fig:method} illustrates the cross-attention maps before and after the application of cross-condition attention modulation. It's observable that compared to \textit{before modulation}, each sub-instruction becomes more concentrated within its respective area of interest.

\subsection{Mask Guided Disentangle Sample}
\label{sec:disentangle_sample}

While restricting the area of interest at the cross-attention level is useful, it is insufficient for fine-grained editing due to the low resolution of semantically rich layers in cross attention~\cite{attend,attention,xie2023boxdiff}. Additionally, disentangling the different noise estimates in~\cref{eq:modified.score.estimate}
is challenging, leading to a lack of robustness in the arbitrary selection of $s_I$ and $s_T$.
Therefore, we suggest modifying the noise estimation to isolate the editing area from irrelevant regions during the sampling process and disentangle the directions of editing and preserving the original image.
To achieve this, we first combine the masks corresponding to all sub-instructions to obtain $\mathcal{M}_{union}$:
\begin{equation}
    \mathcal{M}_{union} = Upsample(\bigvee_{i} \mathcal{M}_{e_i})
\end{equation}
where $Upsample$ denotes the operation of upsampling the mask to match the resolution of the latent space. Following this, new score estimates are formulated:
\begin{align}
    \label{eq:modified.modift_score.estimate}
    \tilde{\epsilon}_\theta (&z_t, t, I, T) = \textcolor{teal}{\epsilon_\theta(z_t, t, \emptyset_I, \emptyset_T)} \notag \\ 
                                & + s_I \big( \textcolor{blue}{\epsilon_\theta(z_t, t, I, \emptyset_T)} - \textcolor{teal}{\epsilon_\theta(z_t, t, \emptyset_I, \emptyset_T)}\big) \notag \\
                                & + s_T \big(\textcolor{red}{\epsilon_\theta(z_t, t, I, T)} - \textcolor{blue}{\epsilon_\theta(z_t, t, I, \emptyset_T)}\big) \odot \mathcal{M}_{union}
\end{align}

We refer to the sampling using the above score estimates as disentangle sampling.
In practice, we employ the disentangle sampling for the initial $75\%$ steps,  and switch to the standard IP2P sampling for the remaining $25\%$,.
\section{Experiments}
\label{sec:experiments}

\begin{figure*}[t!]
    \centering
    \begin{tabular}{c}
    \includegraphics[width=1.0\linewidth]{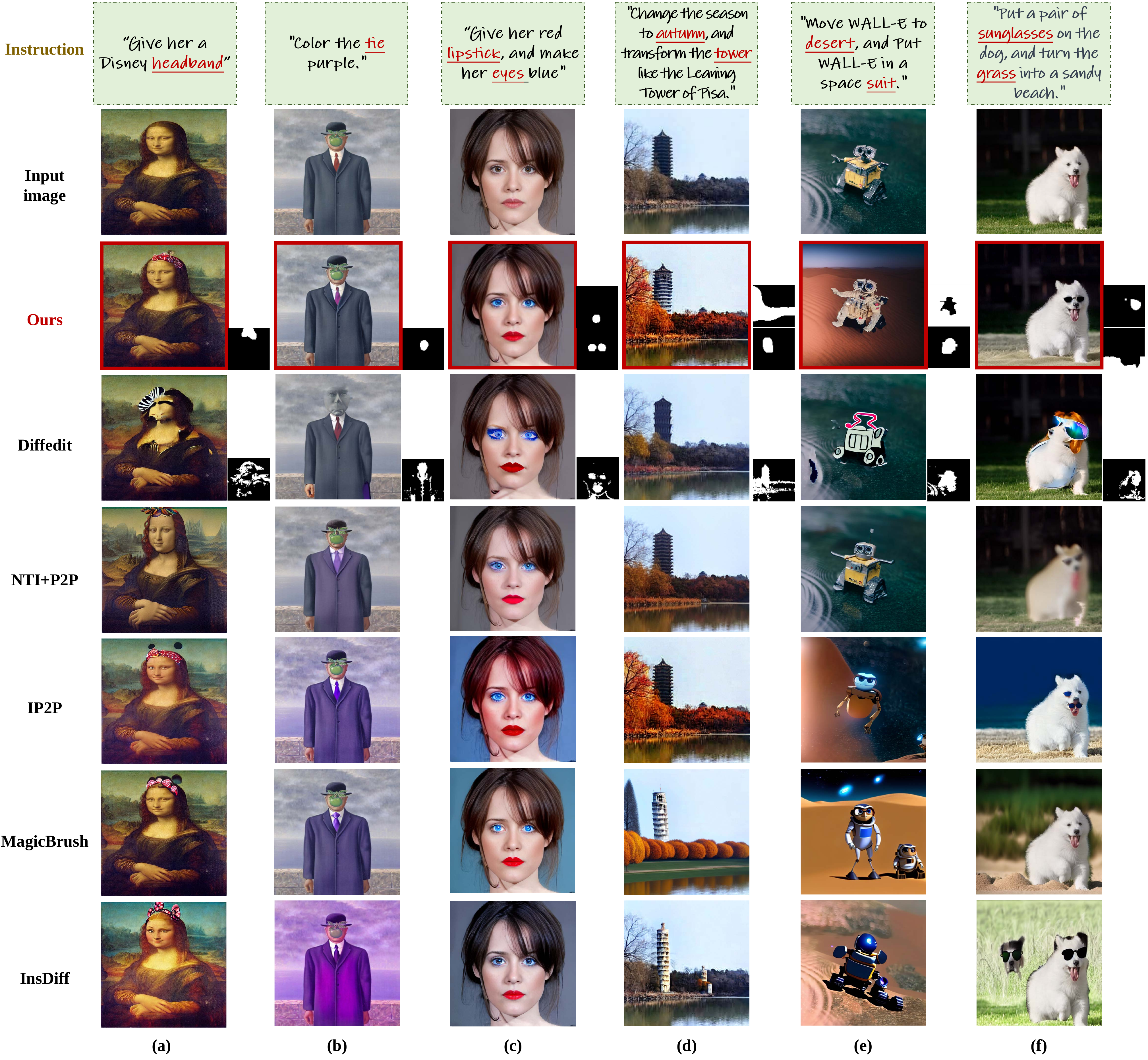} \\
    \end{tabular}
  \caption{
  \textbf{Qualitative comparisons.} We provide all baselines with their desired input formats. From top to bottom: input image, our method, Diffedit~\cite{diffedit}, NTI+P2P~\cite{nulltext}, IP2P~\cite{instructpix2pix}, MagicBrush~\cite{zhang2023magicbrush}, InsDiff~\cite{instructdiffusion}. The texts at the top of the images represent edit instructions. Inputs for Diffedit and NTI+P2P include the original and target captions. Additionally, we present masks that highlight the regions of interest identified by FoI and Diffedit, located on the lower right side of the results, organized in accordance with the sequence of sub-instructions. Compared with baseline models, FoI can accurately edit regions of interest.
  }
  \label{fig:qualitative}
\end{figure*}

\subsection{Experimental Settings}
\label{sec:setting}
\noindent\textbf{Dataset.}
For single-instruction editing, we filter 5,000 localized edit-type images from the IP2P dataset~\cite{instructpix2pix} using GPT4~\cite{gpt4}, each tagged with specific object edits. For multi-instruction editing, we gather 100 real images. For each image, GPT-4V(ision)~\cite{gpt4,gpt4v,gpt4vblog} is used to create 2-4 instructions, along with original and target descriptions, and marking the objects to edit.

\noindent\textbf{Metrics.}
For evaluation, we use \textit{CLIP image similarity}~\cite{clip} and \textit{Dinov2 image similarity}~\cite{oquab2023dinov2} to measure the cosine similarity between edited and original images. \textit{CLIP text-image direction similarity}~\cite{StyleGAN_NADA} evaluates how image changes correspond with changes in their captions. Additionally, \textit{PickScore}~\cite{pickscore} evaluates image fidelity based on learned human preferences.

\noindent\textbf{Baseline models.}
We make comparisons with the state-of-the-art (SOTA) image editing methods, including Diffedit~\cite{diffedit}, NTI+P2P~\cite{nulltext}, IP2P~\cite{instructpix2pix}, MagicBrush~\cite{zhang2023magicbrush}, and InstructDiffusion (InsDiff)~\cite{instructdiffusion}. 
Diffedit identifies regions for editing based on the differences between the noise predictions of original and target prompts. NTI+P2P extends P2P~\cite{hertz2022prompt} to realize the editing of real images, representing inversion-based image editing methods. IP2P is the basic model of instruction-based editing methods. MagicBrush fine-tunes IP2P on its own high-quality constructed dataset. InsDiff uses the same model structure as IP2P but trains a generalist model on multiple datasets.
The closely related Watch Your Steps~\cite{mirzaei2023watch} lacks an available implementation, precluding direct comparison.

\noindent\textbf{Implementation details.}
In all our experiments, we utilize the pretrained IP2P model~\cite{instructpix2pix} with freeze weight.
% available from the diffusers package~\cite{diffusers}.
We use the Euler ancestral sampler~\cite{karras2022elucidating} with a total of 100 denoising steps. The default settings of $s_I=1.5$ and $s_T=7.5$ are used unless specified otherwise.
For mask extraction described in~\cref{sec:get_mask}, it is only performed  during the first denoising step. The threshold \( \tau \) is randomly sampled from the range \([0.4,0.7]\), and the hyperparameter \(\gamma\) is set to 3. Following previous work~\cite{mirzaei2023watch}, the initial noise is generated by adding $80\%$ noise to the original image,
thus the actual number of denoising steps is 80 in FoI.

\subsection{Main Results}
\noindent\textbf{Qualitative evaluation.}
We show some qualitative experimental results in~\cref{fig:qualitative}, From our experiments, we observe the following: Firstly, Diffedit and NTI+P2P often overlook sub-tasks in complex editing scenarios. For example, in~\cref{fig:qualitative} (d), the tower edit is ignored; in~\cref{fig:qualitative} (e), WALL-E remains in water instead of desert; and in~\cref{fig:qualitative} (f), the grass does not change to a sandy bench. These models also lead to unwanted modifications within the editing areas, such as over-modification of Mona Lisa's facial features (\cref{fig:qualitative} (a)), incorrect facial edits (\cref{fig:qualitative} (b)), and excessive changes to the dog in~\cref{fig:qualitative} (f). Secondly, instruction-based methods like IP2P, MagicBrush, and InsDiff tend to over-edit. This is observable as IP2P and MagicBrush modify beyond the intended headband area in~\cref{fig:qualitative} (a), affecting Mona Lisa’s identity. In~\cref{fig:qualitative} (b), they alter the entire image to purple and in~\cref{fig:qualitative} (c), they change the background to blue. Over-editing is also evident in~\cref{fig:qualitative} (d) and (e) by MagicBrush and InsDiff, while IP2P misses the tower edit in~\cref{fig:qualitative} (d). All three methods cause excessive edits to WALL-E in~\cref{fig:qualitative} (e), leading to a significant departure from the original image. IP2P and MagicBrush fail to appropriately add sunglasses to the dog in~\cref{fig:qualitative} (f), and InsDiff creates an incomplete additional dog head wearing sunglasses. Despite the instruction only requesting the glass to change to a sandy bench, IP2P, MagicBrush, and InsDiff also modify the background.

Compared to Diffedit, our method extracts masks for the area of interest with greater precision and detail, ensuring higher quality and more fine-grained editing. 
This further evidences the implicit grounding capability of IP2P in enhancing fine-grained editing.

Our method provides more nuanced editing capabilities and does not affect unrelated areas, outperforming baseline models, especially in scenarios with multi-instruction. 

\begin{figure*}[!t]
    \centering
    \begin{tabular}{c}
    \includegraphics[width=\linewidth]{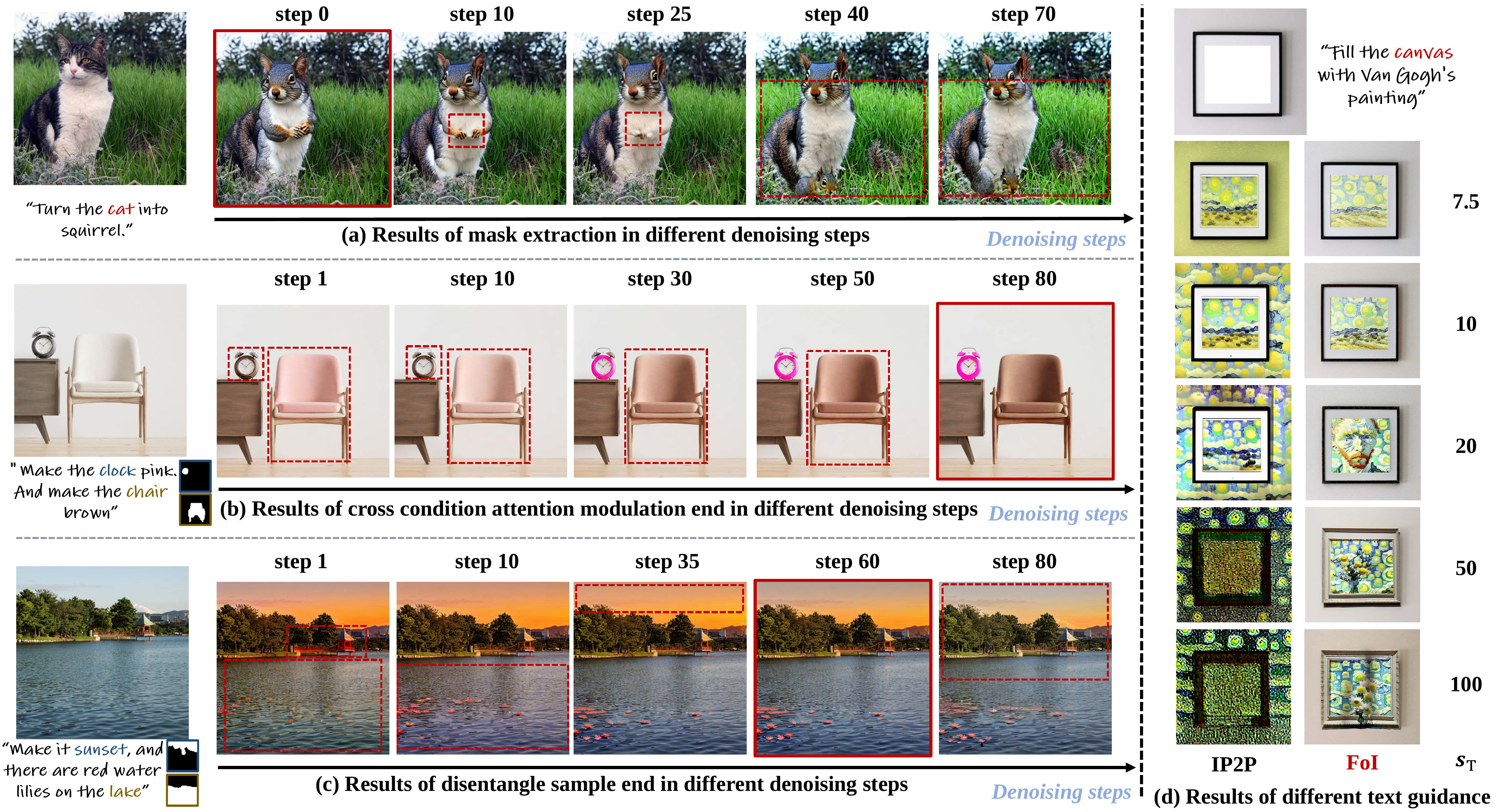} \\
    \end{tabular}
  \caption{\textbf{Ablation study of different components.} The editing effect within the red box is poor. \textbf{(a)} Editing results of mask extraction at different denoising steps. \textbf{(b)} Results of ending cross condition attention modulation at different denoising steps. \textbf{(c)} Results of ending disentangle sampling at different denoising steps. \textbf{(d)}. Robustness of our method compared to IP2P with fixed \(s_I\) across various \(s_T\) values. For (b) and (c), the mask extracted for the editing area are displayed to the right of each instruction.}
  \vspace{-12pt}
  \label{fig:ablation}
\end{figure*}

\begin{table}[!t]
\centering
\setlength{\tabcolsep}{5pt}
\resizebox{\linewidth}{!}{
\begin{tabular}{ll c c  c c }
\toprule
& \textbf{Method}
& \shortstack[c]{\textbf{CLIP-I} } 
& \shortstack[c]{\textbf{Dino-I} }
& \shortstack[c]{\textbf{CLIP-D}}
& \shortstack[c]{\textbf{PickScore}}\\
\midrule
\multirow{6}{*}{\shortstack[c]{\textbf{Single-}\\ \textbf{Instruction} } }  
& Diffedit~\cite{diffedit}    & 0.8627 & 0.7916 &  0.0844 & 0.0639  \\
& NTI+P2P~\cite{nulltext} & 0.8522   & 0.7928 & 0.0981 & 0.0951 \\
\cdashline{2-6}
& IP2P~\cite{instructpix2pix} & 0.8605 & 0.8264 & 0.1685 & 0.1353  \\
& MagicBrush~\cite{zhang2023magicbrush} & 0.9178 & 0.8702 & 0.1934 & 0.1780  \\
& InsDiff~\cite{instructdiffusion}   & 0.8755 & 0.8612 & \textbf{0.2064} & 0.1377 \\
& \textbf{FoI (ours)}    & \textbf{0.9402} & \textbf{0.9277} & 0.1699 & \textbf{0.3901} \\
\midrule

\multirow{6}{*}{\shortstack[c]{\textbf{Multi-}\\ \textbf{Instruction} } } 
& Diffedit~\cite{diffedit}    & 0.8505 & 0.7529 &  0.0629 & 0.0616  \\
& NTI+P2P~\cite{nulltext} & 0.8560   & 0.7526 & 0.0865 & 0.0332 \\
\cdashline{2-6}
& IP2P~\cite{instructpix2pix} & 0.8769 & 0.8369 & 0.1605 & 0.1059  \\
& MagicBrush~\cite{zhang2023magicbrush} & 0.8609 & 0.8291 & \textbf{0.1807} & 0.1591  \\
& InsDiff~\cite{instructdiffusion}   & 0.8439 & 0.7938 & 0.1785 & 0.1325 \\
& \textbf{FoI (ours)}     & \textbf{0.9255} & \textbf{0.9159} & 0.1685 & \textbf{0.5077} \\
\bottomrule
%\vspace{-10pt}
\end{tabular}
}
\vspace{-8pt}
\caption{\textbf{Quantitative comparisons.} We compare our model with baseline models in terms of CLIP image similarity, Dinov2 image similarity, CLIP direction similarity, and PickScore. Our method achieves state-of-the-art results in image similarity and PickScore. Because our method aims to minimize over-editing, the CLIP direction similarity is lower than MagicBrush~\cite{zhang2023magicbrush} and InsDiff~\cite{instructdiffusion}, which tend to over-edit.}
\label{tbl:results}
%\vspace{-6pt}
\end{table}

\noindent\textbf{Quantitative evaluation.}
As illustrated in~\cref{tbl:results}, in single-instruction and multi-instruction evaluations, we achieve state-of-the-art results in CLIP image similarity, Dinov2 image similarity and PickScore, demonstrating that our method best aligns with human perception in terms of fidelity to the original and edited images. Notably, our method shows vastly improved performance over baseline models for the multi-instruction editing task, demonstrating the superiority of our method when faced with complex editing instructions.

In the CLIP direction similarity metric, our method scores lower than MagicBrush~\cite{zhang2023magicbrush} and InsDiff~\cite{instructdiffusion} because they tend towards over-editing, making larger changes to the input image in the direction of the instructions, whereas our method focuses on necessary edits and minimizes effects on irrelevant areas. CLIP~\cite{clip} itself has difficulty perceiving fine-grained changes~\cite{paiss2022no}, so this also proves that we perform more fine-grained editing. Over-editing would lead to reduced CLIP image similarity and Dinov2 image similarity, and increased CLIP direction similarity. Our method balances the preservation of details in the original image and execution of editing instructions.

\begin{table}[!t]
\centering
\resizebox{\linewidth}{!}{
\begin{tabular}{@{}lccccc@{}}
\toprule
 & \multicolumn{2}{c}{\textbf{Single-Instruction}} & \multicolumn{2}{c}{\textbf{Multi-Instruction}} \\
\cmidrule(lr){2-3} \cmidrule(lr){4-5}
& \textbf{Instruction} & \textbf{Image} & \textbf{Instruction} & \textbf{Image} \\
& \textbf{Align} & \textbf{Align} & \textbf{Align} & \textbf{Align} \\
\midrule
Diffedit~\cite{diffedit} & 9.42\% & 10.42\% & 0.75\% & 3.08\% \\
NTI+P2P~\cite{nulltext}      & 9.5\% & 16.58\% & 0.42\% & 4.25\% \\
IP2P~\cite{instructpix2pix}     & 12.83\% & 10.92\% & 3.08\% & 3.92\% \\
MagicBrush~\cite{zhang2023magicbrush}     & \textbf{23.17\%} & 22.83\% & 9.42\% & 4.75\% \\
InsDiff~\cite{instructdiffusion}     & 21.92\% & 11.75\% & 5.50\% & 2.67\% \\
\textbf{FoI (ours)}      & \textbf{23.17\%} & \textbf{27.5\%} & \textbf{80.83\%} & \textbf{81.33\%} \\
\bottomrule
\end{tabular}
}
\vspace{-8pt}
\caption{\textbf{Human preference study.} FoI outperforms baseline models in both instruction- and image-alignment, and achieves a huge advantage in multi-instruction measures. }
\vspace{-0.2cm}
\label{tbl:human_eval}
\end{table}

\noindent\textbf{Human Preference Study.}
For single and multi instruction edits, we conduct a human preference study using 20 images for each category, comparing our FoI with Diffedit, NTI+P2P, IP2P, MagicBrush, and InsDiff. The study includes 60 participants. For instruction alignment, participants are asked to choose the method that best matched the editing effect of the instruction. For image alignment, they select the method that best preserved the original image details (i.e., no changes occurred in unrelated areas). As indicated in~\cref{tbl:human_eval}, our FoI method is favored over the baseline methods for both single and multi instruction edits, with a significant preference gap observed in the multi-instruction editing scenarios, over 80\% of participants perceive the editing quality and fidelity of FoI to surpass that of the baseline models. This underscores the superiority of FoI in precise and high-quality editing tasks.

More results are available in~\cref{supply:addition_res}.

\subsection{Ablation Study}
\label{sec:ablation}
\noindent\textbf{Mask Extraction Steps.}
As illustrated in~\cref{fig:ablation} (a), searching for the mask over extended time steps does not enhance the outcomes; rather, it results in the inadvertent editing of unrelated areas in more denoising steps, and also diminishes the effectiveness within the intended editing regions.

\noindent\textbf{Cross Condition Attention Modulation.}
As shown in~\cref{fig:ablation} (b), halting cross-attention modulation at various denoising steps can have different effects. Early termination might cause instructions to affect unrelated areas, particularly in multi-instruction scenarios, where it can also disrupt other instructions. Notably, the effectiveness of sub-instructions increases with the number of steps conducted.

\noindent\textbf{Disentangle Sample.}
As illustrated in~\cref{fig:ablation} (c), even with the application of cross-attention modulation, the use of broad adjectives can inadvertently result in minor modifications to irrelevant areas.
The Disentangle Sample method alleviates the issue of insufficient granularity in attention modulation, effectively separating the editing areas from irrelevant regions.
However, using disentangle sampling for all steps can lead to suboptimal outcomes. For instance, at \textit{step 80} in~\cref{fig:ablation} (c), the \textit{``Make it sunset.''} instruction creates a more fragmented effect compared to the smoother result at \textit{step 60}.
Moreover, as demonstrated in~\cref{fig:ablation} (d), where we set $s_I=1.5$ and progressively increase $s_T$, unlike previous methods~\cite{instructpix2pix,mirzaei2023watch,joseph2023iterative,haque2023instruct} that require precise tuning of the balance between $s_I$ and $s_T$, our approach maintains this balance with greater robustness.

Quantitative evaluation and analysis from our ablation study will be detailed in~\cref{supply:ablation}.

% \textbf{\textit{Quantitative evaluation and analysis from our ablation study will be detailed in the supplementary materials.}}
\section{Conclusion}
\label{sec:conclusion}
We propose FoI, a tuning-free method that empowers the pretrained IP2P model to execute precise single-instruction edits as well as multi-instruction edits. We discover the IP2P model's implicit grounding capability and extract masks corresponding to each instruction. Furthermore, we utilize these masks for cross-condition attention modulation, which confines instructions within their respective masks while reducing interference between different instructions. Finally, we introduce disentangle sampling, designed to isolate editing areas from irrelevant regions and disentangle the directions of editing and preserving the original image. Our approach demonstrates exceptional performance in both qualitative and quantitative experiments.

\noindent\textbf{Limitations.}
Our approach encounters certain limitations. Although smaller cross-attention maps are rich in semantic content, they restrict our ultra-fine editing ability to an extent. Furthermore, our method's effectiveness is heavily dependent on the capabilities of the pretrained IP2P model.

{
    \small
    \bibliographystyle{ieeenat_fullname}
    \bibliography{main}

\begin{thebibliography}{67}
\providecommand{\natexlab}[1]{#1}
\providecommand{\url}[1]{\texttt{#1}}
\expandafter\ifx\csname urlstyle\endcsname\relax
  \providecommand{\doi}[1]{doi: #1}\else
  \providecommand{\doi}{doi: \begingroup \urlstyle{rm}\Url}\fi

\bibitem[gpt(2023)]{gpt4vblog}
Chatgpt can now see, hear, and speak.
\newblock \url{https://openai.com/blog/chatgpt-can-now-see-hear-and-speak}, 2023.

\bibitem[Agarwal et~al.(2023)Agarwal, Karanam, Joseph, Saxena, Goswami, and Srinivasan]{Astar}
Aishwarya Agarwal, Srikrishna Karanam, KJ Joseph, Apoorv Saxena, Koustava Goswami, and Balaji~Vasan Srinivasan.
\newblock A-star: Test-time attention segregation and retention for text-to-image synthesis.
\newblock \emph{arXiv preprint arXiv:2306.14544}, 2023.

\bibitem[Avrahami et~al.(2022)Avrahami, Lischinski, and Fried]{blend_diffusion}
Omri Avrahami, Dani Lischinski, and Ohad Fried.
\newblock Blended diffusion for text-driven editing of natural images.
\newblock In \emph{Proceedings of the IEEE/CVF Conference on Computer Vision and Pattern Recognition (CVPR)}, pages 18208--18218, 2022.

\bibitem[Avrahami et~al.(2023)Avrahami, Fried, and Lischinski]{blend_diffusion_latent}
Omri Avrahami, Ohad Fried, and Dani Lischinski.
\newblock Blended latent diffusion.
\newblock \emph{ACM Transactions on Graphics (TOG)}, 42\penalty0 (4):\penalty0 1--11, 2023.

\bibitem[Balaji et~al.(2022)Balaji, Nah, Huang, Vahdat, Song, Kreis, Aittala, Aila, Laine, Catanzaro, et~al.]{balaji2022ediffi}
Yogesh Balaji, Seungjun Nah, Xun Huang, Arash Vahdat, Jiaming Song, Karsten Kreis, Miika Aittala, Timo Aila, Samuli Laine, Bryan Catanzaro, et~al.
\newblock ediffi: Text-to-image diffusion models with an ensemble of expert denoisers.
\newblock \emph{arXiv preprint arXiv:2211.01324}, 2022.

\bibitem[Bar-Tal et~al.(2022)Bar-Tal, Ofri-Amar, Fridman, Kasten, and Dekel]{text2live}
Omer Bar-Tal, Dolev Ofri-Amar, Rafail Fridman, Yoni Kasten, and Tali Dekel.
\newblock Text2live: Text-driven layered image and video editing.
\newblock In \emph{European conference on computer vision}, pages 707--723. Springer, 2022.

\bibitem[Brooks et~al.(2023)Brooks, Holynski, and Efros]{instructpix2pix}
Tim Brooks, Aleksander Holynski, and Alexei~A. Efros.
\newblock Instructpix2pix: Learning to follow image editing instructions.
\newblock In \emph{arXiv}, 2023.

\bibitem[Brown et~al.(2020)Brown, Mann, Ryder, Subbiah, Kaplan, Dhariwal, Neelakantan, Shyam, Sastry, Askell, et~al.]{brown2020language}
Tom Brown, Benjamin Mann, Nick Ryder, Melanie Subbiah, Jared~D Kaplan, Prafulla Dhariwal, Arvind Neelakantan, Pranav Shyam, Girish Sastry, Amanda Askell, et~al.
\newblock Language models are few-shot learners.
\newblock \emph{Advances in neural information processing systems}, 33:\penalty0 1877--1901, 2020.

\bibitem[Chefer et~al.(2023)Chefer, Alaluf, Vinker, Wolf, and Cohen-Or]{attend}
Hila Chefer, Yuval Alaluf, Yael Vinker, Lior Wolf, and Daniel Cohen-Or.
\newblock Attend-and-excite: Attention-based semantic guidance for text-to-image diffusion models.
\newblock \emph{ACM Transactions on Graphics (TOG)}, 42\penalty0 (4):\penalty0 1--10, 2023.

\bibitem[Couairon et~al.(2022)Couairon, Verbeek, Schwenk, and Cord]{diffedit}
Guillaume Couairon, Jakob Verbeek, Holger Schwenk, and Matthieu Cord.
\newblock Diffedit: Diffusion-based semantic image editing with mask guidance.
\newblock \emph{arXiv preprint arXiv:2210.11427}, 2022.

\bibitem[Fu et~al.(2023)Fu, Hu, Du, Wang, Yang, and Gan]{instruct_guiding}
Tsu-Jui Fu, Wenze Hu, Xianzhi Du, William~Yang Wang, Yinfei Yang, and Zhe Gan.
\newblock Guiding instruction-based image editing via multimodal large language models.
\newblock \emph{arXiv preprint arXiv:2309.17102}, 2023.

\bibitem[Gal et~al.(2022{\natexlab{a}})Gal, Alaluf, Atzmon, Patashnik, Bermano, Chechik, and Cohen-Or]{gal2022image}
Rinon Gal, Yuval Alaluf, Yuval Atzmon, Or Patashnik, Amit~H Bermano, Gal Chechik, and Daniel Cohen-Or.
\newblock An image is worth one word: Personalizing text-to-image generation using textual inversion.
\newblock \emph{arXiv preprint arXiv:2208.01618}, 2022{\natexlab{a}}.

\bibitem[Gal et~al.(2022{\natexlab{b}})Gal, Patashnik, Maron, Bermano, Chechik, and Cohen-Or]{StyleGAN_NADA}
Rinon Gal, Or Patashnik, Haggai Maron, Amit~H Bermano, Gal Chechik, and Daniel Cohen-Or.
\newblock Stylegan-nada: Clip-guided domain adaptation of image generators.
\newblock \emph{ACM Transactions on Graphics (TOG)}, 41\penalty0 (4):\penalty0 1--13, 2022{\natexlab{b}}.

\bibitem[Geng et~al.(2023)Geng, Yang, Hang, Li, Gu, Zhang, Bao, Zhang, Hu, Chen, et~al.]{instructdiffusion}
Zigang Geng, Binxin Yang, Tiankai Hang, Chen Li, Shuyang Gu, Ting Zhang, Jianmin Bao, Zheng Zhang, Han Hu, Dong Chen, et~al.
\newblock Instructdiffusion: A generalist modeling interface for vision tasks.
\newblock \emph{arXiv preprint arXiv:2309.03895}, 2023.

\bibitem[Goodfellow et~al.(2020)Goodfellow, Pouget-Abadie, Mirza, Xu, Warde-Farley, Ozair, Courville, and Bengio]{GANs}
Ian Goodfellow, Jean Pouget-Abadie, Mehdi Mirza, Bing Xu, David Warde-Farley, Sherjil Ozair, Aaron Courville, and Yoshua Bengio.
\newblock Generative adversarial networks.
\newblock \emph{Communications of the ACM}, 63\penalty0 (11):\penalty0 139--144, 2020.

\bibitem[Haque et~al.(2023)Haque, Tancik, Efros, Holynski, and Kanazawa]{haque2023instruct}
Ayaan Haque, Matthew Tancik, Alexei Efros, Aleksander Holynski, and Angjoo Kanazawa.
\newblock Instruct-nerf2nerf: Editing 3d scenes with instructions.
\newblock In \emph{Proceedings of the IEEE/CVF International Conference on Computer Vision}, 2023.

\bibitem[Hertz et~al.(2022)Hertz, Mokady, Tenenbaum, Aberman, Pritch, and Cohen-Or]{hertz2022prompt}
Amir Hertz, Ron Mokady, Jay Tenenbaum, Kfir Aberman, Yael Pritch, and Daniel Cohen-Or.
\newblock Prompt-to-prompt image editing with cross attention control.
\newblock \emph{arXiv preprint arXiv:2208.01626}, 2022.

\bibitem[Ho and Salimans(2022)]{ho2022classifier}
Jonathan Ho and Tim Salimans.
\newblock Classifier-free diffusion guidance.
\newblock \emph{arXiv preprint arXiv:2207.12598}, 2022.

\bibitem[Ho et~al.(2020)Ho, Jain, and Abbeel]{ho2020denoising}
Jonathan Ho, Ajay Jain, and Pieter Abbeel.
\newblock Denoising diffusion probabilistic models.
\newblock In \emph{arXiv}, 2020.

\bibitem[Hou et~al.(2022)Hou, Shen, Patashnik, Cohen-Or, and Huang]{hou2022feat}
Xianxu Hou, Linlin Shen, Or Patashnik, Daniel Cohen-Or, and Hui Huang.
\newblock Feat: Face editing with attention.
\newblock \emph{arXiv preprint arXiv:2202.02713}, 2022.

\bibitem[Joseph et~al.(2023)Joseph, Udhayanan, Shukla, Agarwal, Karanam, Goswami, and Srinivasan]{joseph2023iterative}
KJ Joseph, Prateksha Udhayanan, Tripti Shukla, Aishwarya Agarwal, Srikrishna Karanam, Koustava Goswami, and Balaji~Vasan Srinivasan.
\newblock Iterative multi-granular image editing using diffusion models.
\newblock \emph{arXiv preprint arXiv:2309.00613}, 2023.

\bibitem[Karazija et~al.(2023)Karazija, Laina, Vedaldi, and Rupprecht]{karazija2023diffusion}
Laurynas Karazija, Iro Laina, Andrea Vedaldi, and Christian Rupprecht.
\newblock Diffusion models for zero-shot open-vocabulary segmentation.
\newblock \emph{arXiv preprint arXiv:2306.09316}, 2023.

\bibitem[Karras et~al.(2022)Karras, Aittala, Aila, and Laine]{karras2022elucidating}
Tero Karras, Miika Aittala, Timo Aila, and Samuli Laine.
\newblock Elucidating the design space of diffusion-based generative models.
\newblock \emph{Advances in Neural Information Processing Systems}, 35:\penalty0 26565--26577, 2022.

\bibitem[Kawar et~al.(2023)Kawar, Zada, Lang, Tov, Chang, Dekel, Mosseri, and Irani]{imagic}
Bahjat Kawar, Shiran Zada, Oran Lang, Omer Tov, Huiwen Chang, Tali Dekel, Inbar Mosseri, and Michal Irani.
\newblock Imagic: Text-based real image editing with diffusion models.
\newblock In \emph{Proceedings of the IEEE/CVF Conference on Computer Vision and Pattern Recognition}, pages 6007--6017, 2023.

\bibitem[Kirillov et~al.(2023)Kirillov, Mintun, Ravi, Mao, Rolland, Gustafson, Xiao, Whitehead, Berg, Lo, et~al.]{sam}
Alexander Kirillov, Eric Mintun, Nikhila Ravi, Hanzi Mao, Chloe Rolland, Laura Gustafson, Tete Xiao, Spencer Whitehead, Alexander~C Berg, Wan-Yen Lo, et~al.
\newblock Segment anything.
\newblock \emph{arXiv preprint arXiv:2304.02643}, 2023.

\bibitem[Kirstain et~al.(2023)Kirstain, Polyak, Singer, Matiana, Penna, and Levy]{pickscore}
Yuval Kirstain, Adam Polyak, Uriel Singer, Shahbuland Matiana, Joe Penna, and Omer Levy.
\newblock Pick-a-pic: An open dataset of user preferences for text-to-image generation.
\newblock \emph{arXiv preprint arXiv:2305.01569}, 2023.

\bibitem[Liu et~al.(2022)Liu, Li, Du, Torralba, and Tenenbaum]{liu2022compositional}
Nan Liu, Shuang Li, Yilun Du, Antonio Torralba, and Joshua~B Tenenbaum.
\newblock Compositional visual generation with composable diffusion models.
\newblock In \emph{European Conference on Computer Vision}, pages 423--439. Springer, 2022.

\bibitem[Liu et~al.(2023)Liu, Zeng, Ren, Li, Zhang, Yang, Li, Yang, Su, Zhu, et~al.]{grounding_dino}
Shilong Liu, Zhaoyang Zeng, Tianhe Ren, Feng Li, Hao Zhang, Jie Yang, Chunyuan Li, Jianwei Yang, Hang Su, Jun Zhu, et~al.
\newblock Grounding dino: Marrying dino with grounded pre-training for open-set object detection.
\newblock \emph{arXiv preprint arXiv:2303.05499}, 2023.

\bibitem[Lyu et~al.(2023)Lyu, Lin, Li, He, Dong, and Tan]{deltaedit}
Yueming Lyu, Tianwei Lin, Fu Li, Dongliang He, Jing Dong, and Tieniu Tan.
\newblock Deltaedit: Exploring text-free training for text-driven image manipulation.
\newblock In \emph{Proceedings of the IEEE Conference on Computer Vision and Pattern Recognition (CVPR)}, 2023.

\bibitem[Meng et~al.(2021)Meng, He, Song, Song, Wu, Zhu, and Ermon]{sdedit}
Chenlin Meng, Yutong He, Yang Song, Jiaming Song, Jiajun Wu, Jun-Yan Zhu, and Stefano Ermon.
\newblock Sdedit: Guided image synthesis and editing with stochastic differential equations.
\newblock \emph{arXiv preprint arXiv:2108.01073}, 2021.

\bibitem[Mirzaei et~al.(2023)Mirzaei, Aumentado-Armstrong, Brubaker, Kelly, Levinshtein, Derpanis, and Gilitschenski]{mirzaei2023watch}
Ashkan Mirzaei, Tristan Aumentado-Armstrong, Marcus~A Brubaker, Jonathan Kelly, Alex Levinshtein, Konstantinos~G Derpanis, and Igor Gilitschenski.
\newblock Watch your steps: Local image and scene editing by text instructions.
\newblock \emph{arXiv preprint arXiv:2308.08947}, 2023.

\bibitem[Mokady et~al.(2023)Mokady, Hertz, Aberman, Pritch, and Cohen-Or]{nulltext}
Ron Mokady, Amir Hertz, Kfir Aberman, Yael Pritch, and Daniel Cohen-Or.
\newblock Null-text inversion for editing real images using guided diffusion models.
\newblock In \emph{Proceedings of the IEEE/CVF Conference on Computer Vision and Pattern Recognition}, pages 6038--6047, 2023.

\bibitem[Nichol et~al.(2021)Nichol, Dhariwal, Ramesh, Shyam, Mishkin, McGrew, Sutskever, and Chen]{nichol2021glide}
Alex Nichol, Prafulla Dhariwal, Aditya Ramesh, Pranav Shyam, Pamela Mishkin, Bob McGrew, Ilya Sutskever, and Mark Chen.
\newblock Glide: Towards photorealistic image generation and editing with text-guided diffusion models.
\newblock \emph{arXiv preprint arXiv:2112.10741}, 2021.

\bibitem[OpenAI(2023{\natexlab{a}})]{gpt4}
OpenAI.
\newblock Gpt-4 technical report, 2023{\natexlab{a}}.

\bibitem[OpenAI(2023{\natexlab{b}})]{gpt4v}
OpenAI.
\newblock Gpt-4v(ision) system card.
\newblock 2023{\natexlab{b}}.

\bibitem[Oquab et~al.(2023)Oquab, Darcet, Moutakanni, Vo, Szafraniec, Khalidov, Fernandez, Haziza, Massa, El-Nouby, et~al.]{oquab2023dinov2}
Maxime Oquab, Timoth{\'e}e Darcet, Th{\'e}o Moutakanni, Huy Vo, Marc Szafraniec, Vasil Khalidov, Pierre Fernandez, Daniel Haziza, Francisco Massa, Alaaeldin El-Nouby, et~al.
\newblock Dinov2: Learning robust visual features without supervision.
\newblock \emph{arXiv preprint arXiv:2304.07193}, 2023.

\bibitem[Paiss et~al.(2022)Paiss, Chefer, and Wolf]{paiss2022no}
Roni Paiss, Hila Chefer, and Lior Wolf.
\newblock No token left behind: Explainability-aided image classification and generation.
\newblock In \emph{European Conference on Computer Vision}, pages 334--350. Springer, 2022.

\bibitem[Parmar et~al.(2023)Parmar, Kumar~Singh, Zhang, Li, Lu, and Zhu]{parmar2023zero}
Gaurav Parmar, Krishna Kumar~Singh, Richard Zhang, Yijun Li, Jingwan Lu, and Jun-Yan Zhu.
\newblock Zero-shot image-to-image translation.
\newblock In \emph{ACM SIGGRAPH 2023 Conference Proceedings}, pages 1--11, 2023.

\bibitem[Patashnik et~al.(2021)Patashnik, Wu, Shechtman, Cohen-Or, and Lischinski]{styleclip}
Or Patashnik, Zongze Wu, Eli Shechtman, Daniel Cohen-Or, and Dani Lischinski.
\newblock Styleclip: Text-driven manipulation of stylegan imagery.
\newblock In \emph{Proceedings of the IEEE/CVF International Conference on Computer Vision (ICCV)}, pages 2085--2094, 2021.

\bibitem[Patashnik et~al.(2023)Patashnik, Garibi, Azuri, Averbuch-Elor, and Cohen-Or]{patashnik2023localizing}
Or Patashnik, Daniel Garibi, Idan Azuri, Hadar Averbuch-Elor, and Daniel Cohen-Or.
\newblock Localizing object-level shape variations with text-to-image diffusion models.
\newblock \emph{arXiv preprint arXiv:2303.11306}, 2023.

\bibitem[Podell et~al.(2023)Podell, English, Lacey, Blattmann, Dockhorn, M{\"u}ller, Penna, and Rombach]{sdxl}
Dustin Podell, Zion English, Kyle Lacey, Andreas Blattmann, Tim Dockhorn, Jonas M{\"u}ller, Joe Penna, and Robin Rombach.
\newblock Sdxl: improving latent diffusion models for high-resolution image synthesis.
\newblock \emph{arXiv preprint arXiv:2307.01952}, 2023.

\bibitem[Radford et~al.(2021)Radford, Kim, Hallacy, Ramesh, Goh, Agarwal, Sastry, Askell, Mishkin, Clark, Krueger, and Sutskever]{clip}
Alec Radford, Jong~Wook Kim, Chris Hallacy, Aditya Ramesh, Gabriel Goh, Sandhini Agarwal, Girish Sastry, Amanda Askell, Pamela Mishkin, Jack Clark, Gretchen Krueger, and Ilya Sutskever.
\newblock Learning transferable visual models from natural language supervision.
\newblock In \emph{ICML}, 2021.

\bibitem[Ramesh et~al.(2022)Ramesh, Dhariwal, Nichol, Chu, and Chen]{ramesh2022hierarchical}
Aditya Ramesh, Prafulla Dhariwal, Alex Nichol, Casey Chu, and Mark Chen.
\newblock Hierarchical text-conditional image generation with clip latents.
\newblock \emph{arXiv preprint arXiv:2204.06125}, 1\penalty0 (2):\penalty0 3, 2022.

\bibitem[Revanur et~al.(2023)Revanur, Basu, Agrawal, Agarwal, and Pai]{revanur2023coralstyleclip}
Ambareesh Revanur, Debraj Basu, Shradha Agrawal, Dhwanit Agarwal, and Deepak Pai.
\newblock Coralstyleclip: Co-optimized region and layer selection for image editing.
\newblock In \emph{Proceedings of the IEEE/CVF Conference on Computer Vision and Pattern Recognition}, pages 12695--12704, 2023.

\bibitem[Rombach et~al.(2022)Rombach, Blattmann, Lorenz, Esser, and Ommer]{latentdiffusion}
Robin Rombach, Andreas Blattmann, Dominik Lorenz, Patrick Esser, and Bj\"orn Ommer.
\newblock High-resolution image synthesis with latent diffusion models.
\newblock In \emph{Proceedings of the IEEE/CVF Conference on Computer Vision and Pattern Recognition (CVPR)}, pages 10684--10695, 2022.

\bibitem[Ruiz et~al.(2023)Ruiz, Li, Jampani, Pritch, Rubinstein, and Aberman]{ruiz2023dreambooth}
Nataniel Ruiz, Yuanzhen Li, Varun Jampani, Yael Pritch, Michael Rubinstein, and Kfir Aberman.
\newblock Dreambooth: Fine tuning text-to-image diffusion models for subject-driven generation.
\newblock In \emph{Proceedings of the IEEE/CVF Conference on Computer Vision and Pattern Recognition}, pages 22500--22510, 2023.

\bibitem[Saharia et~al.(2022)Saharia, Chan, Saxena, Li, Whang, Denton, Ghasemipour, Gontijo~Lopes, Karagol~Ayan, Salimans, et~al.]{Imagen}
Chitwan Saharia, William Chan, Saurabh Saxena, Lala Li, Jay Whang, Emily~L Denton, Kamyar Ghasemipour, Raphael Gontijo~Lopes, Burcu Karagol~Ayan, Tim Salimans, et~al.
\newblock Photorealistic text-to-image diffusion models with deep language understanding.
\newblock \emph{Advances in Neural Information Processing Systems}, 35:\penalty0 36479--36494, 2022.

\bibitem[Schuhmann et~al.(2022)Schuhmann, Beaumont, Vencu, Gordon, Wightman, Cherti, Coombes, Katta, Mullis, Wortsman, et~al.]{schuhmann2022laion}
Christoph Schuhmann, Romain Beaumont, Richard Vencu, Cade Gordon, Ross Wightman, Mehdi Cherti, Theo Coombes, Aarush Katta, Clayton Mullis, Mitchell Wortsman, et~al.
\newblock Laion-5b: An open large-scale dataset for training next generation image-text models.
\newblock \emph{Advances in Neural Information Processing Systems}, 35:\penalty0 25278--25294, 2022.

\bibitem[Sohl-Dickstein et~al.(2015)Sohl-Dickstein, Weiss, Maheswaranathan, and Ganguli]{sohl2015deep}
Jascha Sohl-Dickstein, Eric Weiss, Niru Maheswaranathan, and Surya Ganguli.
\newblock Deep unsupervised learning using nonequilibrium thermodynamics.
\newblock In \emph{International conference on machine learning}, pages 2256--2265. PMLR, 2015.

\bibitem[Song et~al.(2020)Song, Meng, and Ermon]{DDIM}
Jiaming Song, Chenlin Meng, and Stefano Ermon.
\newblock Denoising diffusion implicit models.
\newblock \emph{arXiv preprint arXiv:2010.02502}, 2020.

\bibitem[Tang et~al.(2022)Tang, Pandey, Jiang, Yang, Kumar, Lin, and Ture]{daam}
Raphael Tang, Akshat Pandey, Zhiying Jiang, Gefei Yang, Karun Kumar, Jimmy Lin, and Ferhan Ture.
\newblock What the daam: Interpreting stable diffusion using cross attention.
\newblock \emph{arXiv preprint arXiv:2210.04885}, 2022.

\bibitem[Tian et~al.(2023)Tian, Aggarwal, Colaco, Kira, and Gonzalez-Franco]{tian2023diffuse}
Junjiao Tian, Lavisha Aggarwal, Andrea Colaco, Zsolt Kira, and Mar Gonzalez-Franco.
\newblock Diffuse, attend, and segment: Unsupervised zero-shot segmentation using stable diffusion.
\newblock \emph{arXiv preprint arXiv:2308.12469}, 2023.

\bibitem[Tumanyan et~al.(2023)Tumanyan, Geyer, Bagon, and Dekel]{pnp}
Narek Tumanyan, Michal Geyer, Shai Bagon, and Tali Dekel.
\newblock Plug-and-play diffusion features for text-driven image-to-image translation.
\newblock In \emph{Proceedings of the IEEE/CVF Conference on Computer Vision and Pattern Recognition}, pages 1921--1930, 2023.

\bibitem[Van Den~Oord et~al.(2017)Van Den~Oord, Vinyals, et~al.]{van2017neural}
Aaron Van Den~Oord, Oriol Vinyals, et~al.
\newblock Neural discrete representation learning.
\newblock \emph{Advances in neural information processing systems}, 30, 2017.

\bibitem[Vaswani et~al.(2017)Vaswani, Shazeer, Parmar, Uszkoreit, Jones, Gomez, Kaiser, and Polosukhin]{attention}
Ashish Vaswani, Noam Shazeer, Niki Parmar, Jakob Uszkoreit, Llion Jones, Aidan~N Gomez, {\L}ukasz Kaiser, and Illia Polosukhin.
\newblock Attention is all you need.
\newblock In \emph{NeurIPS}, 2017.

\bibitem[Wallace et~al.(2023)Wallace, Gokul, and Naik]{wallace2023edict}
Bram Wallace, Akash Gokul, and Nikhil Naik.
\newblock Edict: Exact diffusion inversion via coupled transformations.
\newblock In \emph{Proceedings of the IEEE/CVF Conference on Computer Vision and Pattern Recognition}, pages 22532--22541, 2023.

\bibitem[Wang et~al.(2023{\natexlab{a}})Wang, Yang, Yang, Butt, and van~de Weijer]{dpl}
Kai Wang, Fei Yang, Shiqi Yang, Muhammad~Atif Butt, and Joost van~de Weijer.
\newblock Dynamic prompt learning: Addressing cross-attention leakage for text-based image editing.
\newblock \emph{arXiv preprint arXiv:2309.15664}, 2023{\natexlab{a}}.

\bibitem[Wang et~al.(2023{\natexlab{b}})Wang, Zhang, Birsak, and Wonka]{instructedit}
Qian Wang, Biao Zhang, Michael Birsak, and Peter Wonka.
\newblock Instructedit: Improving automatic masks for diffusion-based image editing with user instructions.
\newblock \emph{arXiv preprint arXiv:2305.18047}, 2023{\natexlab{b}}.

\bibitem[Wu et~al.(2023{\natexlab{a}})Wu, Zhao, Chen, Gu, Zhao, He, Zhou, Shou, and Shen]{wu2023datasetdm}
Weijia Wu, Yuzhong Zhao, Hao Chen, Yuchao Gu, Rui Zhao, Yefei He, Hong Zhou, Mike~Zheng Shou, and Chunhua Shen.
\newblock Datasetdm: Synthesizing data with perception annotations using diffusion models.
\newblock \emph{arXiv preprint arXiv:2308.06160}, 2023{\natexlab{a}}.

\bibitem[Wu et~al.(2023{\natexlab{b}})Wu, Zhao, Shou, Zhou, and Shen]{wu2023diffumask}
Weijia Wu, Yuzhong Zhao, Mike~Zheng Shou, Hong Zhou, and Chunhua Shen.
\newblock Diffumask: Synthesizing images with pixel-level annotations for semantic segmentation using diffusion models.
\newblock \emph{arXiv preprint arXiv:2303.11681}, 2023{\natexlab{b}}.

\bibitem[Xie et~al.(2023)Xie, Li, Huang, Liu, Zhang, Zheng, and Shou]{xie2023boxdiff}
Jinheng Xie, Yuexiang Li, Yawen Huang, Haozhe Liu, Wentian Zhang, Yefeng Zheng, and Mike~Zheng Shou.
\newblock Boxdiff: Text-to-image synthesis with training-free box-constrained diffusion.
\newblock In \emph{Proceedings of the IEEE/CVF International Conference on Computer Vision}, pages 7452--7461, 2023.

\bibitem[Xu et~al.(2022)Xu, Lin, Tang, Li, He, Sebe, Timofte, Van~Gool, and Ding]{xu2022predict}
Zipeng Xu, Tianwei Lin, Hao Tang, Fu Li, Dongliang He, Nicu Sebe, Radu Timofte, Luc Van~Gool, and Errui Ding.
\newblock Predict, prevent, and evaluate: Disentangled text-driven image manipulation empowered by pre-trained vision-language model.
\newblock In \emph{Proceedings of the IEEE/CVF Conference on Computer Vision and Pattern Recognition}, pages 18229--18238, 2022.

\bibitem[Xue et~al.(2023)Xue, Song, Guo, Liu, Zong, Liu, and Luo]{xue2023raphael}
Zeyue Xue, Guanglu Song, Qiushan Guo, Boxiao Liu, Zhuofan Zong, Yu Liu, and Ping Luo.
\newblock Raphael: Text-to-image generation via large mixture of diffusion paths.
\newblock \emph{arXiv preprint arXiv:2305.18295}, 2023.

\bibitem[Yang et~al.(2023)Yang, Gui, Wang, Chen, Zhuang, and Shen]{objectaware}
Zhen Yang, Dinggang Gui, Wen Wang, Hao Chen, Bohan Zhuang, and Chunhua Shen.
\newblock Object-aware inversion and reassembly for image editing, 2023.

\bibitem[Zhang et~al.(2023{\natexlab{a}})Zhang, Mo, Chen, Sun, and Su]{zhang2023magicbrush}
Kai Zhang, Lingbo Mo, Wenhu Chen, Huan Sun, and Yu Su.
\newblock Magicbrush: A manually annotated dataset for instruction-guided image editing.
\newblock \emph{arXiv preprint arXiv:2306.10012}, 2023{\natexlab{a}}.

\bibitem[Zhang et~al.(2023{\natexlab{b}})Zhang, Rao, and Agrawala]{zhang2023adding}
Lvmin Zhang, Anyi Rao, and Maneesh Agrawala.
\newblock Adding conditional control to text-to-image diffusion models.
\newblock In \emph{Proceedings of the IEEE/CVF International Conference on Computer Vision}, pages 3836--3847, 2023{\natexlab{b}}.

\bibitem[Zhang et~al.(2023{\natexlab{c}})Zhang, Yang, Feng, Qin, Chen, Yu, Chen, Wang, Savarese, Ermon, et~al.]{hive}
Shu Zhang, Xinyi Yang, Yihao Feng, Can Qin, Chia-Chih Chen, Ning Yu, Zeyuan Chen, Huan Wang, Silvio Savarese, Stefano Ermon, et~al.
\newblock Hive: Harnessing human feedback for instructional visual editing.
\newblock \emph{arXiv preprint arXiv:2303.09618}, 2023{\natexlab{c}}.

\end{thebibliography}
}

% WARNING: do not forget to delete the supplementary pages from your submission 
% \input{sec/X_suppl}
% \clearpage

\appendix

\maketitlesupplementary

\begin{figure*}[!t]
    \centering
    \begin{tabular}{c}
    \includegraphics[width=\linewidth]{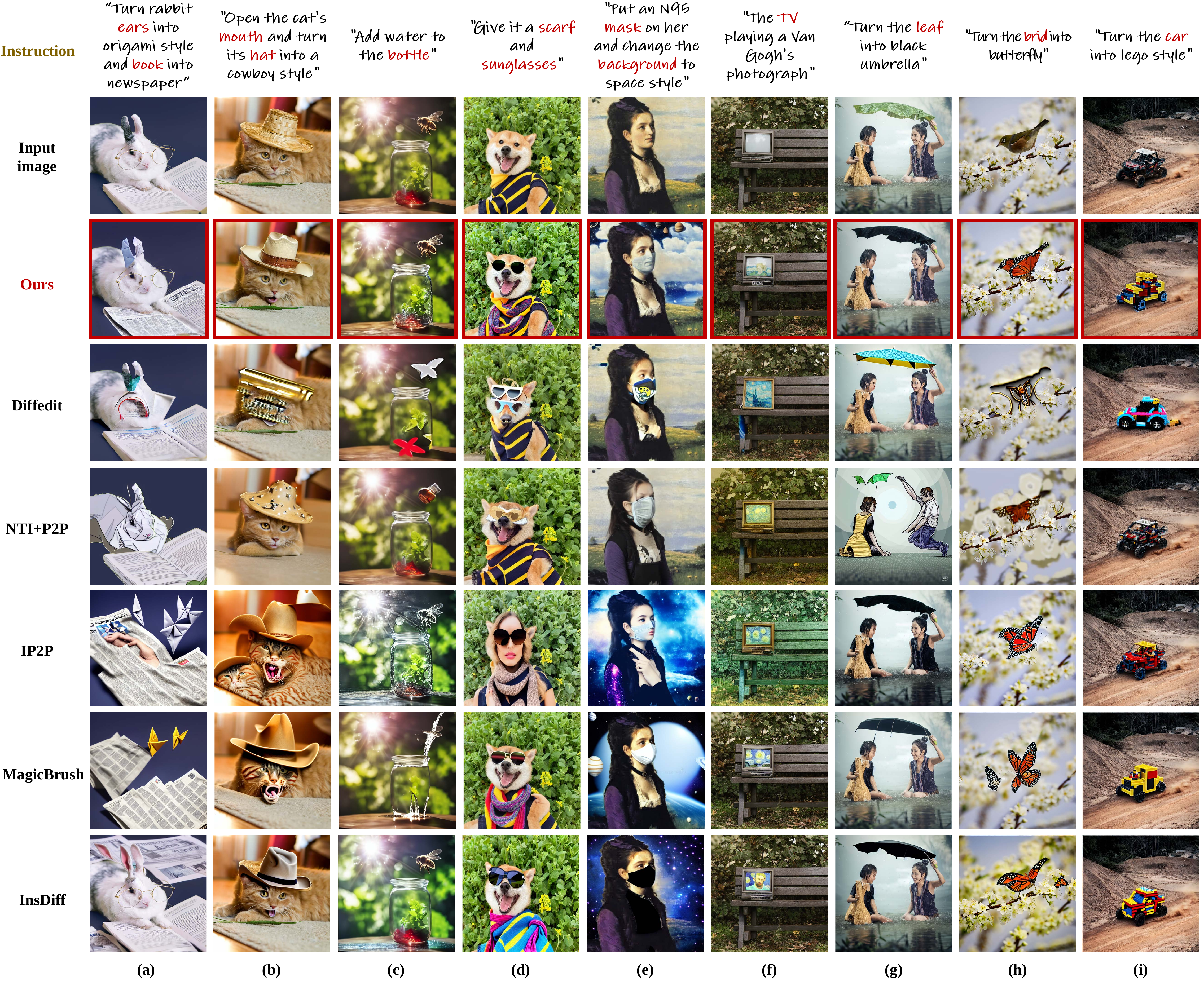} \\
    \end{tabular}
  \caption{\textbf{Additional qualitative comparison.}}
  % \vspace{-6pt}
  \label{fig:addition_qua}
\end{figure*}

\section{Ablation Study}
\label{supply:ablation}
In this section, we offer a detailed quantitative analysis of each component within our method. We compiled a dataset of 40 real images, each paired with 1 to 4 instructions. Our analysis primarily revolves around the changes in \textit{CLIP image similarity} (hereafter referred to as \textit{CLIP-I}) and \textit{CLIP text-image direction similarity} (hereafter referred to as \textit{CLIP-D}). The former metric gauges image similarity, while the latter assesses the degree to which the editing direction is followed.

\noindent\textbf{Mask Extraction Steps.}
\cref{table:mask-extraction} illustrates the impact of mask extraction at various denoising steps. Notably, after completing mask extraction, we proceed with standard cross-condition attention modulation and disentangle sampling. It is evident that earlier mask extraction leads to higher scores in both \textit{CLIP-I} and \textit{CLIP-D}.

\noindent\textbf{Cross Condition Attention Modulation.}
\cref{table:cross-condition-attention} shows the effects of ceasing cross-condition attention modulation at different denoising steps. It is important to note that the mask is always extracted in the first denoising step, and disentangle sampling is utilized in $75\%$ of the remaining inference steps. Ceasing at step 1 implies no use of cross-condition attention modulation, whereas ceasing at step 80 means its constant usage. The results show that both \textit{CLIP-I} and \textit{CLIP-D} scores increase with more steps of cross-condition attention modulation, validating its contribution to more granular and precise editing.

\noindent\textbf{Disentangle Sample.}
\cref{table:disentangle-sample} reveals the outcomes of stopping disentangle sampling at different steps. Here, the mask is consistently extracted in the first denoising step, and cross-condition attention modulation is applied in all remaining steps. Stopping at step 1 indicates no use of disentangle sampling, whereas stopping at step 80 represents its continuous use. We observe that with increasing steps of disentangle sampling, \textit{CLIP-I} continually rises, but \textit{CLIP-D} first increases and then decreases. This is due to the suboptimal results when disentangle sampling is used throughout all steps, resulting in inconsistencies across the image. The qualitative outcomes of this are detailed in \cref{sec:ablation}.

\begin{table}[H]
\centering
\resizebox{0.8\linewidth}{!}{
\begin{tabular}{@{}cccccc@{}}
\toprule
\textbf{Step} & 0 & 20 & 40 & 60 & 79 \\
\midrule
\textbf{CLIP-I} & \textbf{0.9260} & 0.9183 & 0.9059 & 0.8910 & 0.8805 \\
\textbf{CLIP-D} & \textbf{0.1745} & 0.1724 & 0.1715 & 0.1708 & 0.1697 \\
\bottomrule
\end{tabular}
}
\caption{Mask Extraction in Different Denoising Steps.}
\label{table:mask-extraction}
\end{table}

\begin{table}[H]
\centering
\resizebox{0.8\linewidth}{!}{
\begin{tabular}{@{}cccccc@{}}
\toprule
\textbf{Step} & 1 & 20 & 40 & 60 & 80 \\
\midrule
\textbf{CLIP-I} & 0.9172 & 0.9179 & 0.9201 & 0.9248 & \textbf{0.9260} \\
\textbf{CLIP-D} & 0.1701 & 0.1715 & 0.1729 & 0.1736 & \textbf{0.1745} \\
\bottomrule
\end{tabular}
}
\caption{Cross condition attention modulation end in different denoising steps.}
\label{table:cross-condition-attention}
\end{table}

\begin{table}[H]
\centering
\resizebox{0.8\linewidth}{!}{
\begin{tabular}{@{}cccccc@{}}
\toprule
\textbf{Step} & 1 & 20 & 40 & 60 & 80 \\
\midrule
\textbf{CLIP-I} & 0.9176 & 0.9185 & 0.9190 & 0.9260 & \textbf{0.9329} \\
\textbf{CLIP-D} & 0.1729 & 0.1732 & 0.1738 & \textbf{0.1745} & 0.1729 \\
\bottomrule
\end{tabular}
}
\caption{Disentangle sample end in different denoising steps.}
\label{table:disentangle-sample}
\end{table}

\section{Additional Results}
\label{supply:addition_res}

\subsection{Additional Qualitative Results}
\label{supply:addition_quali_res}
In~\cref{fig:addition_qua}, we provide additional qualitative comparisons between our method and baseline models.

\subsection{Enhanced Control Over Sub-Instruction Intensity}
\label{supply:sub_int}
\begin{figure*}[!t]
    \centering
    \begin{tabular}{c}
    \includegraphics[width=\linewidth]{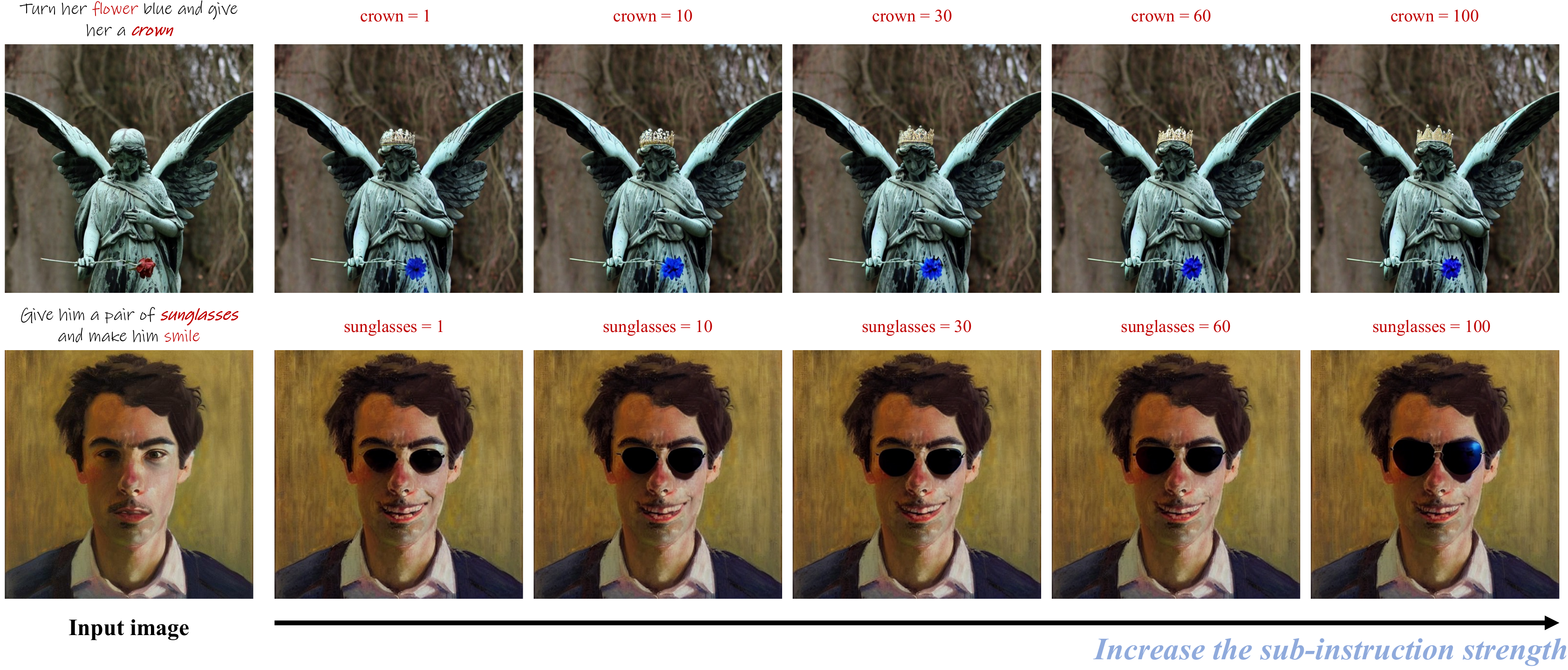} \\
    \end{tabular}
  \caption{\textbf{Controling sub-instruction guidance scale.} Every result image is labeled with the enhancement scale of the corresponding sub-instruction.}
  % \vspace{-6pt}
  \label{fig:sub_ins}
\end{figure*}

As mentioned in~\cref{sec:attention_modulation}, as illustrated in~\cref{fig:sub_ins}, our method enables flexible control over the intensity of different sub-instructions, providing a level of finesse and granularity not achievable with previous methods.

\subsection{Comparison of Editing Speed}
\label{supply:speed}
The evaluation of diverse editing techniques' speed was conducted on a GeForce RTX 3090, with results detailed in~\cref{table:time}. This assessment involved randomly selecting 200 images and calculating the average inference time. To guarantee a fair comparison, the inference steps for each model were standardized at 50 steps. Consequently, for FoI, the effective number of inference steps utilized is 40.

\begin{table}[H]
\centering
\resizebox{\linewidth}{!}{
\begin{tabular}{@{}ccccccc@{}}
\toprule
\textbf{Method} & Diffedit & NTI+P2P & IP2P & MagicBrush & InsDiff & \textbf{FoI(ours)} \\
\midrule
\textbf{Avg Time Cost(s)} & 15.60 & 138.20 & 7.35 & 7.34 & 7.60 & \textbf{6.79} \\
\bottomrule
\end{tabular}
}
\caption{Inference time comparison.}
\label{table:time}
\end{table}

\subsection{Human Preference Study}
\begin{figure*}[!t]
    \centering
    \begin{tabular}{c}
    \includegraphics[width=\linewidth]{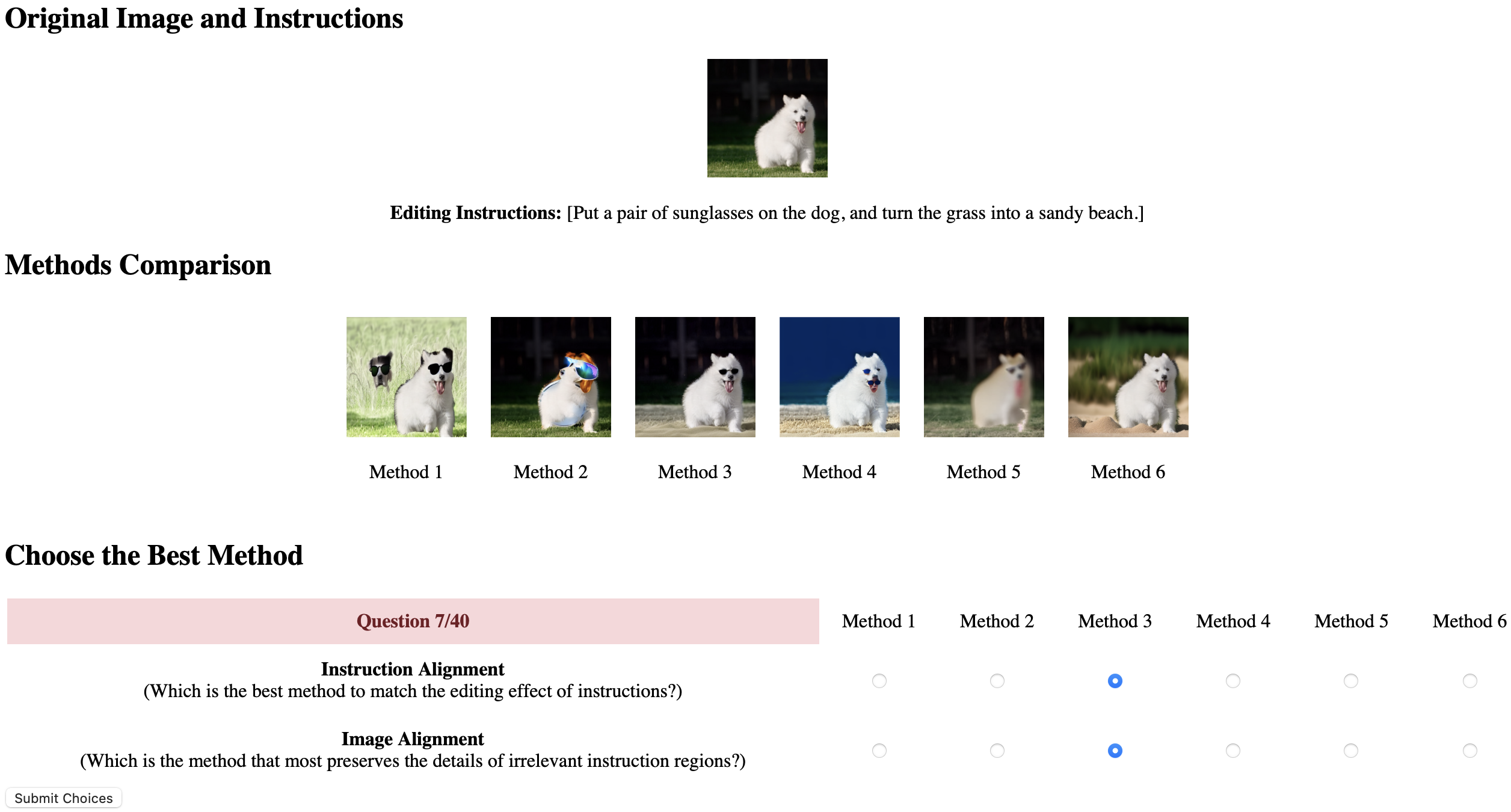} \\
    \end{tabular}
  \caption{\textbf{Human preference study print screen.}}
  % \vspace{-6pt}
  \label{fig:user_study}
\end{figure*}

~\cref{fig:user_study} presents the questionnaire form used in our human preference study, where the display order of the six methods was randomized for each question.

\section{Societal Impact}
Our study introduces a fine-grained, multi-instruction editing scheme for images. This nuanced alteration of images could potentially be exploited by malevolent entities to produce false content and disseminate misinformation, a well-known issue inherent to all image editing techniques. However, our method uniquely generates a mask for the edited regions upon completion, aiding in the training of models to detect forged images. We believe our work contributes significantly to this field by providing an analysis of instruction-based image editing methods.

\end{document}